\documentclass[11pt]{article}
\usepackage[final]{acl}
\usepackage{times}
\usepackage{latexsym}
\usepackage[T1]{fontenc}
\usepackage[utf8]{inputenc}
\usepackage{microtype}
\usepackage{inconsolata}
\usepackage{graphicx}
\usepackage{amsmath,amssymb}
\usepackage{booktabs}
\usepackage{colortbl}
\usepackage{arydshln}
\usepackage{multirow}
\usepackage{algorithm}
\usepackage{algorithmic}
\usepackage{subcaption}
\usepackage{enumitem}
\usepackage{amsfonts}
\usepackage{caption}
\usepackage{xcolor} 
\PassOptionsToPackage{table}{xcolor}

\definecolor{ao}{rgb}{0.0, 0.0, 1.0}
\usepackage[most]{tcolorbox} 

\usepackage{enumitem}

\usepackage{hyperref}

\newtcolorbox{boxK}{
    sharpish corners, 
    boxrule = 0pt,
    toprule = 4.5pt, 
    enhanced,
    fuzzy shadow = {0pt}{-2pt}{-0.5pt}{0.5pt}{black!35} 
}

\newcommand{\sota}[1]{\textcolor{ao}{\textbf{#1}}}

\title{
    \texttt{s3}: You Don't Need That Much Data to Train a Search Agent via RL

}
\author{Pengcheng Jiang,\; Xueqiang Xu,\; Jiacheng Lin,\; Jinfeng Xiao\footnotemark[2]$^2$,\\  \textbf{Zifeng Wang}, \textbf{Jimeng Sun}, and \textbf{Jiawei Han} \\\\
University of Illinois Urbana Champaign \quad \textsuperscript{\dag}Amazon\\
  \texttt{\{pj20,jimeng,hanj\}@illinois.edu} \quad \texttt{jfx@amazon.com} \\
  }
  
\begin{document}
\maketitle
\footnotetext[1]{Code available at \url{https://github.com/pat-jj/s3}.}
\footnotetext[2]{Prior to co-author’s role at Amazon.}
\begin{abstract}
Retrieval-augmented generation (RAG) systems empower large language models (LLMs) to access external knowledge during inference. Recent advances have enabled LLMs to act as search agents via reinforcement learning (RL), improving information acquisition through multi-turn interactions with retrieval engines. However, existing approaches either optimize retrieval using search-only metrics (e.g., NDCG) that ignore downstream utility or fine-tune the entire LLM to jointly reason and retrieve—entangling retrieval with generation and limiting the real search utility and compatibility with frozen or proprietary models. In this work, we propose \texttt{s3}, a lightweight, model-agnostic framework that decouples the searcher from the generator and trains the searcher using a Gain Beyond RAG reward: the improvement in generation accuracy over naïve RAG. \texttt{s3} requires only 2.4k training samples to outperform baselines trained on over 70$\times$ more data, consistently delivering stronger downstream performance across six general QA and five medical QA benchmarks.$^\text{1}$
\end{abstract}

\section{Introduction}
\begin{figure*}[t]
\vspace{-1em}
    \centering
    \includegraphics[width=\linewidth]{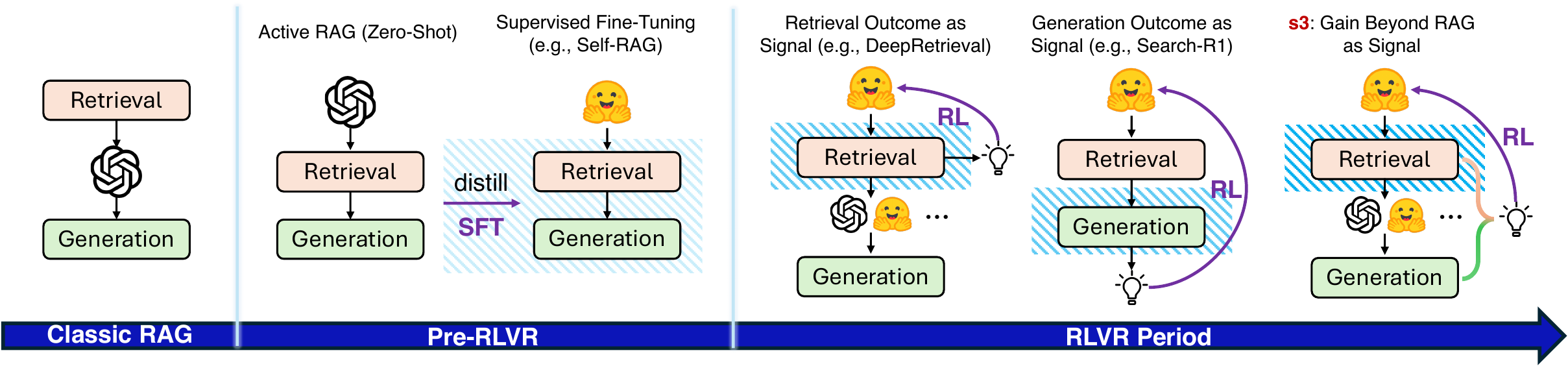}
    \caption{\small RAG has progressed from fixed or supervised retrieval to RL-based agentic methods. While prior work trains retrieval or generation jointly, \texttt{s3} focuses solely on the searcher, improving generation without tuning the generator LLM.}
    \label{fig:rag_progress}
\end{figure*}

Retrieval-Augmented Generation (RAG) enables large language models (LLMs) to access and reason over external knowledge by retrieving relevant documents and conditioning generation on them~\cite{lewis2020retrieval}. 
As shown in Figure~\ref{fig:rag_progress}, we categorize the evolution of RAG systems into three phases.

\begin{figure}[t]
    \vspace{-1em}
    \centering
    \includegraphics[width=\linewidth]{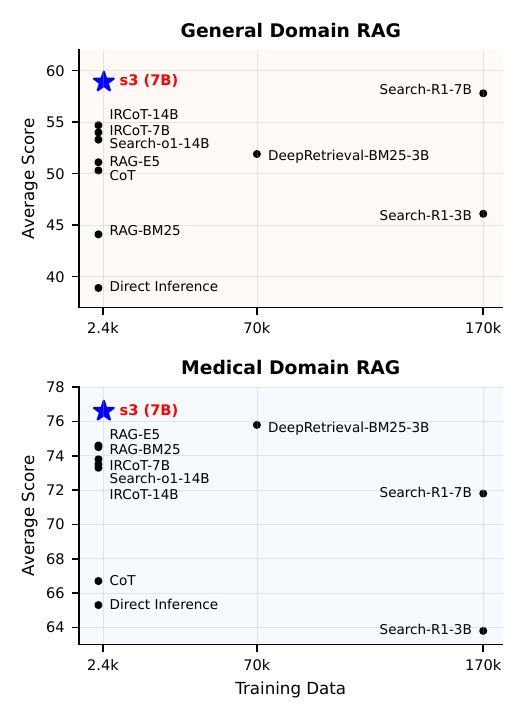}
    \vspace{-2.5em}
    \caption{\textbf{Training Data vs Averaged Performance across six general and five medical QA Datasets} (tested with Claude-3-Haiku as the generator LLM).}
    \label{fig:train_perf_abs}
    \vspace{-1em}
\end{figure}
\begin{figure}[t]
    \vspace{-1em}
    \centering
    \includegraphics[width=\linewidth]{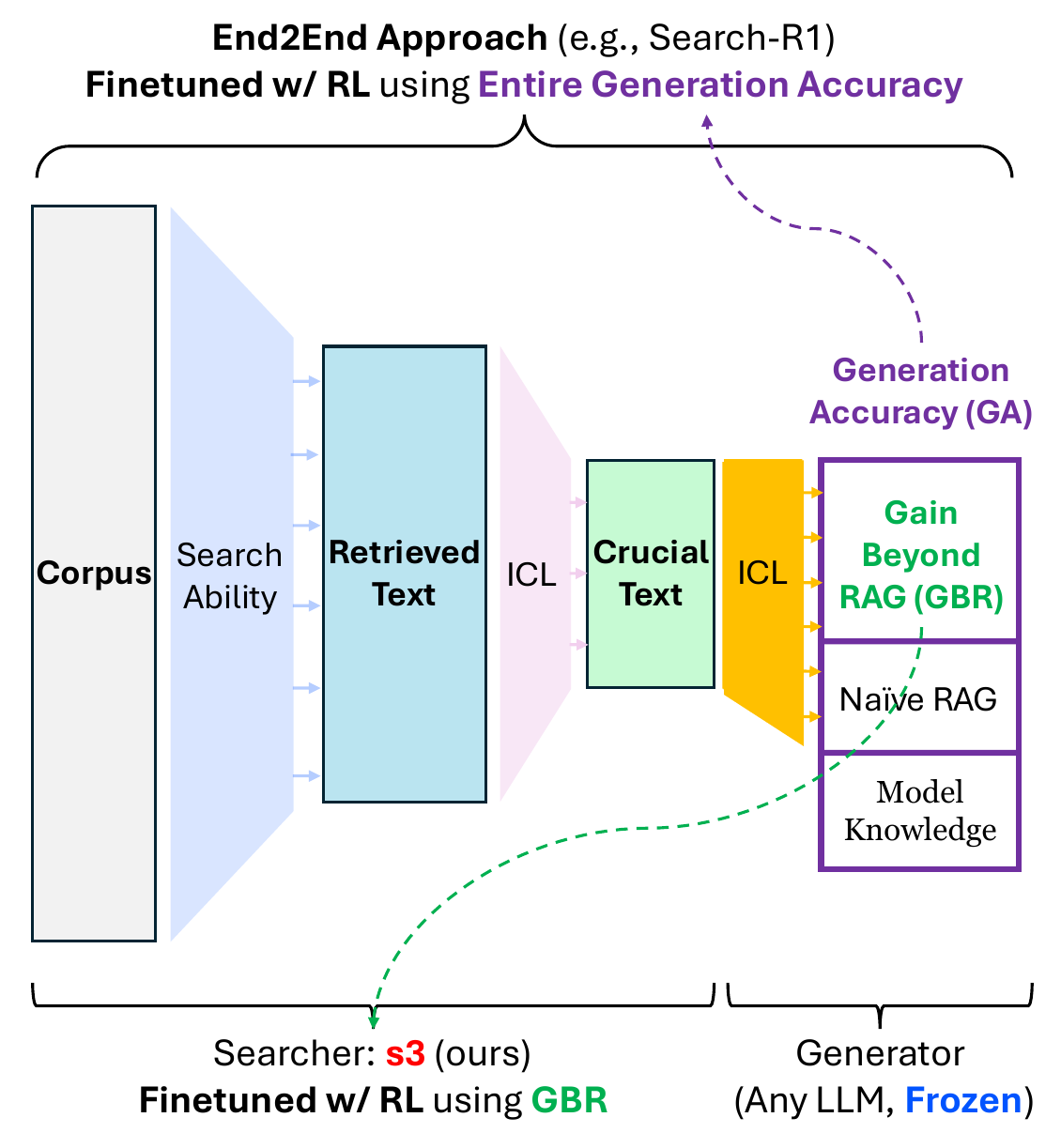}
    \caption{\small \textbf{Decomposition of Agentic RAG.} End-to-end approaches fine-tune the entire model using the entire generation accuracy, making it difficult to isolate the contribution of search. In contrast, \texttt{s3} freezes the generator and trains only the searcher with Gain Beyond RAG (GBR), a novel reward that quantifies the added value of retrieved context over naïve RAG, enabling modular, efficient optimization.}
    \label{fig:decompose}
    \vspace{-0.5em}
\end{figure}

\noindent\textbf{Classic RAG.} Early approaches relied on static retrieval methods, where queries were fixed and retrieval quality was decoupled from downstream generation performance. Despite their simplicity, these systems often underperformed on queries that need contextual or multi-hop reasoning.

\noindent\textbf{Pre-RLVR.} To improve retrieval quality, subsequent methods enabled more active participation of the LLM during inference. Active RAG techniques~\cite{yao2022react,jiang2023active,trivedi2023interleaving} interleaved query generation, retrieval, and reasoning in a multi-turn loop. These systems introduced iterative retrieval but typically relied on zero-shot prompting and lacked trainable components. Self-RAG~\cite{asai2023self} distilled such behaviors from larger models into smaller ones via supervised fine-tuning, teaching smaller models to reason and retrieve effectively without external rewards. While these methods improved flexibility and reduced supervision cost, they still did not optimize retrieval using outcome signals.

\noindent\textbf{RLVR Period.} The recent emergence of reinforcement learning with verifiable rewards (RLVR)~\cite{su2025crossing} marks a new phase. DeepSeek-R1-Zero~\cite{guo2025deepseek} showed that even rule-based, outcome-driven rewards (e.g., answer correctness) can train strong reasoning agents. Building on this idea, DeepRetrieval~\cite{jiang2025deepretrieval} applied RL to train query generators using search-oriented metrics like recall and NDCG. However, these metrics are disconnected from downstream answer quality. Search-R1~\cite{jin2025search} trained a single model to jointly retrieve and generate via reinforcement learning, using exact match (EM) as the reward. While this approach improves answer accuracy, the tight entanglement between search and generation makes it difficult to isolate genuine retrieval improvements (see Figure~\ref{fig:decompose}). Moreover, EM is a brittle reward signal—failing to reward semantically correct answers phrased differently.

This motivates a shift toward a modular framework where search and generation are cleanly separated, and optimization focuses purely on search quality with respect to downstream utility~\cite{dai2025seper}. We propose \texttt{s3}, a simple yet powerful framework that trains a search-only agent using a \emph{novel} reward signal: \textit{Gain Beyond RAG} (GBR). GBR 
measures how much better the generator performs when conditioned on retrieved documents from \texttt{s3}, compared to na{i}ve top-$k$ retrieval. This setup keeps the generator LLM frozen, sidesteps answer token overfitting, and directly optimizes the retrieval component to serve any black-box LLM.

Remarkably, \texttt{s3} achieves strong gains with only 2.4k training examples, outperforming DeepRetrieval (focused on retrieval metrics) and Search-R1 (entangled optimization) both in terms of context quality and final answer performance.

\vspace{0.5em}
\noindent\textbf{Our main contributions are:}
\begin{itemize}[leftmargin=*]
\item We introduce \texttt{s3}, a modular, RL-based search framework that optimizes for generation quality without touching the generator.
\item We define Gain Beyond RAG (GBR), a principled, model-agnostic reward signal that quantifies improvements over standard retrieval.
\item We show that \texttt{s3} outperforms state-of-the-art agentic RAG methods on six general and five medical QA benchmarks, using 70$\times$ less training data (see Figure~\ref{fig:train_perf_abs}).
\end{itemize}

\begin{figure*}[t]
    \centering
    \includegraphics[width=\linewidth]{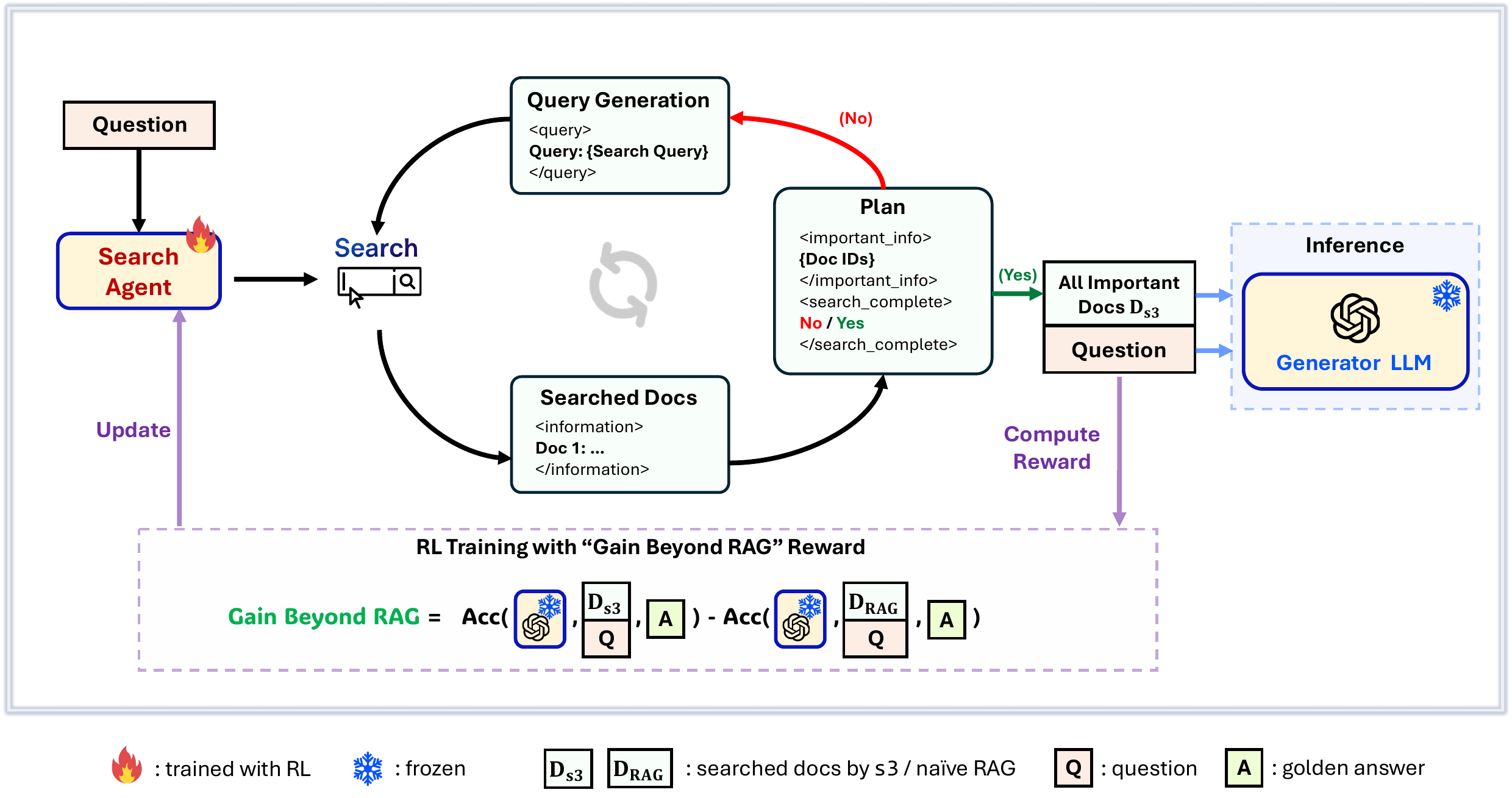}
\caption{\small
\textbf{Overview of the \texttt{s3} framework.}
The search agent iteratively retrieves documents, selects useful documents, and generates queries until completion. The final context $D_{s3}$ is then passed to a frozen generator LLM. The search agent is trained using Gain Beyond RAG (GBR), which quantifies improvement over naïve top-$k$ retrieval from the original question.
}

    \label{fig:framework}
\end{figure*}
\section{Related Work}

\subsection{Retrieval-Augmented Generation}

Large language models (LLMs) have shown impressive generative capabilities~\cite{touvron2023llama2,openai2023gpt4}, but their factuality remains bounded~\cite{peng2023check} by their training corpora. Retrieval-Augmented Generation (RAG)~\cite{lewis2020retrieval,gao2023retrieval} augments LLMs by prepending retrieved documents to their input, enabling access to up-to-date or domain-specific information. The effectiveness of this setup, however, depends heavily on the retrieval quality.
\
Early efforts improve retrieval through supervised query rewriting~\cite{nogueira2019passage,lin2023train}, where LLMs are fine-tuned to generate better search queries from manually labeled or distilled training data. These methods require significant annotation effort and often optimize for imitation rather than end-task performance. Recent works have introduced Active RAG methods~\cite{yao2022react,trivedi2023interleaving,asai2023self,lyu2024retrieve}, which prompt LLMs to iteratively retrieve and reason in a zero-shot or few-shot manner. While flexible, these methods typically rely on handcrafted prompting patterns and lack direct optimization by interacting with environment. 

\subsection{RL for Agentic Retrieval and Search-Centric Optimization}

The emergence of reinforcement learning (RL) for large language models has given rise to agentic retrieval, where models interact with search engines and 
improve by receiving outcome-based feedback—such as whether the final answer is correct.
We refer to this shift as the RL-Zero period, sparked by the insight that even simple rewards like answer correctness can elicit strong reasoning and search behavior~\cite{guo2025deepseek}. 
Within this paradigm, retrieval-centric methods like DeepRetrieval~\cite{jiang2025deepretrieval} optimize query generation for search metrics (e.g., recall, NDCG),
which often fail to reflect answer utility.
Conversely, end-to-end approaches like Search-R1~\cite{jin2025search} train LLMs to retrieve and generate jointly using exact match rewards, but require full model access and entangle search with answer token alignment.

In contrast, \texttt{s3} takes a searcher-centric approach that avoids generator fine-tuning. It directly optimizes retrieval quality using a generation-aware reward, enabling lightweight and modular training that is compatible with black-box LLMs.

\section{\texttt{s3}: Optimized \underline{S}earch-\underline{S}elect-\underline{S}erve Flow with Reinforcement Learning}

We introduce \texttt{s3}, a lightweight, model-agnostic framework that equips a tunable search agent with structured, multi-turn access to external knowledge. As illustrated in Figure~\ref{fig:framework}, the search agent interacts with a search engine iteratively: it generates queries, retrieves documents, selects a subset of useful evidence, and decides whether to continue searching. A frozen generator LLM then consumes the accumulated evidence to produce a final answer.
\
To ensure a fair reward baseline, \texttt{s3} begins by retrieving top-$k$ ($k=3$ in our experiments) documents from the original question, just like naïve RAG. The search agent is trained using the Gain Beyond RAG (GBR) reward, which measures the improvement in generation accuracy when using its retrieved context versus this baseline. This modular design enables targeted optimization of retrieval quality, decoupled from answer generation.

\subsection{Multi-Turn Search-Select Loop}

Given a question $Q$, the system consists of (1) a search agent (LLM policy) $\pi_{\texttt{s3}}$, (2) a search engine $\mathcal{R}$, (3) a frozen generator LLM $\mathcal{G}$.
\texttt{s3} first retrieves top-$k$ documents using $q_0 = Q$, yielding $\mathcal{D}_0 = \mathcal{R}(Q)$. A subset $\mathcal{D}_0^{\text{sel}} \subseteq \mathcal{D}_0$ is selected to form the initial context. It then performs a sequence of search rounds $t = 1, 2, \dots, T$, structured as:

\begin{tcolorbox}[colback=blue!1!white, colframe=blue!20!black, title=\texttt{s3} Loop, sharp corners=south,fonttitle=\bfseries]
\small
\begin{enumerate}[leftmargin=*]
    \item \textbf{Search:} Documents $\mathcal{D}_t = \mathcal{R}(q_t)$ are retrieved in \texttt{<information>}...\texttt{</information>}
    \item \textbf{Select:} $\pi_{\texttt{s3}}$ specifies useful documents in\\ 
    \texttt{<important\_info>}...\texttt{</important\_info>}, corresponding to subset $\mathcal{D}_t^{\text{sel}} \subseteq \mathcal{D}_t$.
    \item \textbf{Stop decision:} $\pi_{\texttt{s3}}$ indicates to stop or not in\\
    \texttt{<search\_complete>}[$1/0$]\texttt{</search\_complete>}.
    \item \textbf{Query Generation:} $\pi_{\texttt{s3}}$ emits a query $q_{t+1}$ in \texttt{<query>}...\texttt{</query>} and go to the \textbf{Search} step.
\end{enumerate}
\vspace{-0.5em}
\end{tcolorbox}

\noindent The loop continues until \texttt{search\_complete} is True (1) or the turn limit is reached. The final context is
$
\mathcal{D}_{\texttt{s3}} = \bigcup_{t=0}^{T} \mathcal{D}_t^{\text{sel}},
$
which is passed (served) to the generator to produce the final output:
\[
\hat{A} = \mathcal{G}(Q, \mathcal{D}_{\texttt{s3}})
\]

\noindent\textbf{Begin with Search.}
Initializing with $q_0 = Q$ ensures the loop begins with the same context as naïve RAG, making the Gain Beyond RAG reward (introduced below) reflect real search improvements.

\subsection{Training via Gain Beyond RAG (GBR)}

To train $\pi_{\texttt{s3}}$, we frame search as a reinforcement learning problem. The reward signal, \emph{Gain Beyond RAG (GBR)}, quantifies the improvement in generation accuracy over a fixed top-$k$ baseline:
\begin{align}
    \text{GBR}(Q) = \text{Acc}(\mathcal{G}(&Q, \mathcal{D}_{\texttt{s3}}), A) \nonumber\\ 
    &- \text{Acc}(\mathcal{G}(Q, \mathcal{D}_{\text{RAG}}), A)
\end{align}
where $A$ is the gold-standard answer, and $\mathcal{D}_{\text{RAG}} = \mathcal{R}(Q)$ is the top-$k$ retrieval from the original question. $\text{Acc}(\cdot)$ is a task-specific metric, which we instantiate as \emph{Generation Accuracy} (see \S\ref{subsec:metric}) for RAG performance.

This reward ensures the search agent is incentivized to retrieve documents that meaningfully enhance the generator’s output quality, independent of surface-form answer similarity.
To improve training efficiency, we precompute the baseline accuracy term $\text{Acc}(\mathcal{G}(Q, \mathcal{D}_{\text{RAG}}), A)$ and restrict training to examples where it equals 0. This effectively filters out questions already solvable by naïve RAG, allowing \texttt{s3} to focus on harder queries where improved retrieval is essential for generation success.

\subsection{Search Policy Optimization}

We optimize the search policy $\pi_{\texttt{s3}}$ via reinforcement learning using the Gain Beyond RAG (GBR) reward. Each rollout consists of a complete search trajectory: emitted queries, document selections, and a stop decision. Once the final context $\mathcal{D}_{\texttt{s3}}$ is constructed, the generator $\mathcal{G}$ produces an answer, and the GBR reward is computed. The generator remains frozen; gradients are backpropagated only through the search policy.
Our method is agnostic to the specific advantage estimation algorithm. In this work, we use \emph{Proximal Policy Optimization} (PPO)~\cite{schulman2017proximal} due to its strong empirical stability~\cite{jiang2025deepretrieval,jin2025search}.
\noindent The PPO objective is:
\begin{align}
    \mathcal{L}_{\texttt{PPO}}(\theta) = 
    \mathbb{E}_{\tau \sim \pi_\theta} &\Big[
    \sum_{t=1}^{T} \min\Big(
    r_t(\theta) \, \hat{A}_t, \nonumber\\
    &\text{clip}(r_t(\theta), 1{-}\epsilon, 1{+}\epsilon) \, \hat{A}_t
    \Big)
    \Big]
\end{align}
where $r_t(\theta) = \frac{\pi_\theta(a_t | s_t)}{\pi_{\text{old}}(a_t | s_t)}$ is the probability ratio between the current and reference policies, $\hat{A}_t$ is the estimated advantage, and $\epsilon$ is clipping threshold.

\begin{table*}[t]
\vspace{-1em}
\centering
\small
\resizebox{\textwidth}{!}{
\begin{tabular}{lrrccccccl}
\toprule
&&&  \multicolumn{3}{c}{\textbf{Single-Hop}} &\multicolumn{3}{c}{\textbf{Multi-Hop}} \\
\cmidrule(lr){4-6} \cmidrule(lr){7-9} 
\textbf{Methods} &\textbf{Searcher} &\textbf{\#Train} & \textbf{NQ\textsuperscript{\textdagger}} & \textbf{TriviaQA} & \textbf{PopQA} & \textbf{HotpotQA\textsuperscript{\textdagger}} & \textbf{2wiki} & \textbf{Musique} & \textbf{Avg.} \\
\midrule
\#Test Data && &3,610 &11,313 &14,267 &7,405 &12,576 &2,417 \\
\midrule
\rowcolor{red!10}
\multicolumn{10}{c}{{End-to-End Fine-Tuning}} \\
SFT$_\text{Qwen2.5-3B-Inst}$ &- &170k &23.7$_\text{(17.5)}$ &41.6$_\text{(34.3)}$ &18.1$_\text{(14.0)}$ &18.0$_\text{(13.7)}$ &22.1$_\text{(20.8)}$ &5.1$_\text{(2.9)}$ &21.4$_\text{(17.2)}$\\
R1$_\text{Qwen2.5-7B-Inst}$ &- &170k &35.6$_\text{(28.8)}$  &60.2$_\text{(53.4)}$ &22.4$_\text{(20.5)}$ &29.4$_\text{(24.0)}$ &30.0$_\text{(29.1)}$ &10.7$_\text{(7.8)}$ &31.4$_\text{(27.3)}$\\
Search-R1-3B   &(self) 3B &170k &47.0$_\text{(27.9)}$ &65.6$_\text{(46.2)}$ &46.4$_\text{(34.9)}$ &33.5$_\text{(22.1)}$ &28.5$_\text{(24.4)}$ &6.0$_\text{(2.8)}$ &37.8$_\text{(26.4)}$\\
Search-R1-7B   &(self) 7B &170k &56.9$_\text{(48.2)}$ &73.8$_\text{(64.0)}$ &50.6$_\text{(46.8)}$ &54.6$_\text{(43.5)}$ &51.6$_\text{(38.4)}$ &28.5$_\text{(20.6)}$ &52.7$_\text{(43.6)}$\\
\midrule
\rowcolor{blue!10}
\multicolumn{10}{c}{{Generator (Qwen2.5-7b-Instruct) Frozen}} \\
Direct Inference &-  &0 &37.3$_\text{(4.4)}$ &55.1$_\text{(32.9)}$ &19.9$_\text{(8.3)}$ &28.1$_\text{(7.6)}$ &36.9$_\text{(9.1)}$ &10.6$_\text{(1.2)}$ &31.3$_\text{(10.6)}$\\
CoT  &- &0 &37.7$_\text{(10.3)}$ &60.6$_\text{(35.4)}$ &22.2$_\text{(11.3)}$ &31.1$_\text{(13.4)}$  &31.6$_\text{(18.9)}$ &10.6$_\text{(4.2)}$ &32.3$_\text{(15.6)}$\\
\hdashline[2pt/2pt] \\ [-8pt]
RAG$_\text{BM25}$   &- &0 &43.6$_\text{(3.8)}$ &69.8$_\text{(29.7)}$ &34.6$_\text{(12.4)}$ &45.3$_\text{(15.1)}$ &38.5$_\text{(10.3)}$ &11.5$_\text{(1.5)}$ &40.6$_\text{(12.1)}$\\
RAG$_\text{E5}$   &- &0 &62.1$_\text{(5.8)}$ &74.5$_\text{(33.8)}$ &54.5$_\text{(20.3)}$ &46.6$_\text{(13.6)}$ &40.1$_\text{(7.8)}$ &13.0$_\text{(2.0)}$ &48.5$_\text{(13.9)}$\\
IRCoT  &(self) 7B &0 &63.2$_\text{(6.2)}$ &75.6$_\text{(34.3)}$ &54.5$_\text{(19.3)}$ &50.9$_\text{(15.4)}$ &48.7$_\text{(9.6)}$ &16.4$_\text{(2.5)}$ &51.6$_\text{(14.5)}$\\
IRCoT  &14B &0 &63.9$_\text{(6.3)}$  &75.5$_\text{(34.9)}$ &55.5$_\text{(20.3)}$ &52.5$_\text{(16.0)}$ &47.4$_\text{(9.3)}$ &17.2$_\text{(2.7)}$ &52.0$_\text{(14.9)}$\\
Search-R1-3B (Ret)  &3B  &170k &56.6$_\text{(6.6)}$ &68.6$_\text{(32.5)}$ &49.4$_\text{(18.8)}$ &41.5$_\text{(13.6)}$ &33.2$_\text{(7.8)}$ &12.1$_\text{(1.9)}$ &43.6$_\text{(13.5)}$\\
Search-R1-7B (Ret)  &7B   &170k &61.3$_\text{(8.1)}$   &73.7$_\text{(35.9)}$ &51.9$_\text{(20.7)}$ &58.6$_\text{(20.0)}$ &50.8$_\text{(12.2)}$ &\sota{27.6$_\text{(7.1)}$} &54.0$_\text{(17.3)}$\\
\hdashline[2pt/2pt] \\ [-8pt]
\texttt{s3}   &7B &2.4k  &\sota{66.1$_\text{(7.2)}$} &\sota{78.5$_\text{(36.8)}$} &\sota{57.4$_\text{(21.9)}$} &\sota{59.0$_\text{(21.8)}$} &\sota{51.6$_\text{(12.4)}$} &23.9$_\text{(6.1)}$ &\sota{56.1$_\text{(17.7)}$}\\
\midrule
\rowcolor{blue!10}
\multicolumn{10}{c}{{Generator (Qwen2.5-14b-Instruct) Frozen}} \\
Direct Inference  &- &0 &38.8$_\text{(8.2)}$ &62.7$_\text{(39.0)}$ &24.5$_\text{(10.8)}$ &30.2$_\text{(9.5)}$ &38.6$_\text{(7.2)}$ &12.5$_\text{(1.8)}$ &34.5$_\text{(12.8)}$\\
CoT   &- &0 &40.5$_\text{(10.2)}$ &66.2$_\text{(41.6)}$ &24.6$_\text{(13.6)}$ &32.9$_\text{(12.3)}$ &33.2$_\text{(13.8)}$ &12.6$_\text{(5.2)}$ &35.0$_\text{(16.1)}$\\
\hdashline[2pt/2pt] \\ [-8pt]
RAG$_\text{BM25}$   &- &0 &54.8$_\text{(16.4)}$  &76.7$_\text{(44.8)}$ & 41.5$_\text{(22.7)}$ & 50.4$_\text{(18.3)}$ &49.9$_\text{(6.4)}$ &17.7$_\text{(3.1)}$ &48.5$_\text{(18.6)}$\\
RAG$_\text{E5}$    &- &0 &62.4$_\text{(18.7)}$ &77.4$_\text{(50.7)}$ &55.1$_\text{(34.0)}$ &47.4$_\text{(20.9)}$ &44.9$_\text{(10.1)}$ &16.1$_\text{(3.3)}$  &50.6$_\text{(23.0)}$\\
IRCoT  &7B &0 &63.0$_\text{(18.8)}$  &77.7$_\text{(50.1)}$ &56.3$_\text{(33.5)}$ &50.7$_\text{(22.7)}$ &53.2$_\text{(12.4)}$ &17.5$_\text{(4.1)}$ &53.1$_\text{(23.6)}$\\
IRCoT  &(self) 14B &0 &63.9$_\text{(19.2)}$ &78.2$_\text{(51.7)}$ &56.1$_\text{(33.8)}$ &51.6$_\text{(23.7)}$ &54.0$_\text{(12.0)}$ &19.1$_\text{(5.2)}$ &53.8$_\text{(24.3)}$\\
Search-R1-3B (Ret)  &3B &170k &59.2$_\text{(16.5)}$  &75.6$_\text{(47.4)}$ &52.3$_\text{(30.3)}$ & 45.5$_\text{(18.3)}$ &44.0$_\text{(8.3)}$ &16.0$_\text{(2.9)}$ &48.8$_\text{(20.6)}$\\
Search-R1-7B (Ret)  &7B &170k  &63.8$_\text{(18.0)}$ &76.3$_\text{(49.5)}$ &54.6$_\text{(33.3)}$ &56.7$_\text{(25.3)}$ & 56.7$_\text{(11.0)}$ &\sota{30.2$_\text{(9.1)}$} &56.4$_\text{(24.4)}$\\
\hdashline[2pt/2pt] \\ [-8pt]
\texttt{s3}   &7B  &2.4k &\sota{67.2$_\text{(18.3)}$} &\sota{79.5$_\text{(48.9)}$}  &\sota{57.8$_\text{(35.7)}$} &\sota{57.1$_\text{(23.3)}$} &\sota{57.1$_\text{(11.6)}$} &26.7$_\text{(7.8)}$ &\sota{57.6$_\text{(24.3)}$}\\
\midrule
\rowcolor{blue!10}
\multicolumn{10}{c}{{Generator (Claude-3-Haiku) Frozen}} \\

Direct Inference &- &0 &48.1$_\text{(25.7)}$ &76.5$_\text{(64.8)}$ &35.7$_\text{(30.9)}$ &35.5$_\text{(24.2)}$ &28.9$_\text{(24.0)}$ &8.8$_\text{(4.3)}$ &38.9$_\text{(29.0)}$\\
CoT  &- &0 &61.5$_\text{(2.9)}$ &81.0$_\text{(30.0)}$ &43.2$_\text{(9.1)}$ &48.8$_\text{(8.8)}$ &46.2$_\text{(6.8)}$ &21.2$_\text{(2.3)}$ &50.3$_\text{(10.0)}$\\
\hdashline[2pt/2pt] \\ [-8pt]
RAG$_\text{BM25}$   &- &0 &50.5$_\text{(3.8)}$ &75.5$_\text{(28.4)}$ &35.9$_\text{(8.0)}$ &50.2$_\text{(11.4)}$ &40.7$_\text{(8.1)}$ &11.8$_\text{(0.8)}$ &44.1$_\text{(10.1)}$\\
DeepRetrieval$_\text{BM25}$ &3B &70k &64.4$_\text{(3.7)}$ &80.2$_\text{(23.2)}$ &45.5$_\text{(8.2)}$ &54.5$_\text{(10.2)}$ &47.1$_\text{(8.0)}$ &22.2$_\text{(1.7)}$ &52.3$_\text{(8.1)}$\\
RAG$_\text{E5}$   &-  &0 &66.5$_\text{(4.3)}$  &80.7$_\text{(28.9)}$ &55.7$_\text{(8.9)}$ &50.7$_\text{(11.5)}$ &39.2$_\text{(7.8)}$ &14.0$_\text{(1.2)}$ &51.1$_\text{(10.4)}$\\
IRCoT  &7B &0 &68.0$_\text{(4.2)}$ &81.7$_\text{(29.3)}$ &55.5$_\text{(8.9)}$ &54.8$_\text{(11.7)}$ &46.5$_\text{(8.1)}$ &17.4$_\text{(1.6)}$ &54.0$_\text{(10.6)}$\\
IRCoT  &14B &0 &68.3$_\text{(4.2)}$ &81.6$_\text{(29.5)}$ &56.1$_\text{(8.6)}$ &55.5$_\text{(11.9)}$ &47.7$_\text{(8.4)}$ &18.9$_\text{(1.7)}$ &54.7$_\text{(10.7)}$\\
Search-o1  &14B & 0 &67.3$_\text{(4.7)}$ &81.2$_\text{(29.8)}$ &50.2$_\text{(9.3)}$ &58.1$_\text{(12.6)}$ &48.8$_\text{(8.4)}$ &14.2$_\text{(1.2)}$ &53.3$_\text{(11.0)}$\\
Search-R1-3B (Ret)  &3B  &170k &60.7$_\text{(3.3)}$ &74.5$_\text{(24.8)}$  &50.1$_\text{(6.9)}$ &45.7$_\text{(10.0)}$ &33.1$_\text{(7.0)}$ &12.7$_\text{(1.3)}$ &46.1$_\text{(8.9)}$\\
Search-R1-7B (Ret)  &7B &170k &68.1$_\text{(4.1)}$ &80.9$_\text{(25.9)}$ &55.7$_\text{(7.0)}$ &62.0$_\text{(11.2)}$ &51.0$_\text{(7.2)}$ &\sota{29.3$_\text{(3.2)}$} &57.8$_\text{(9.8)}$\\
\hdashline[2pt/2pt] \\ [-8pt]
\texttt{s3}  &7B &2.4k &\sota{70.5$_\text{(3.2)}$} &\sota{84.0$_\text{(24.6)}$} &\sota{57.7$_\text{(5.9)}$} &\sota{62.4$_\text{(11.1)}$} &\sota{52.4$_\text{(8.3)}$} &26.2$_\text{(7.9)}$ &\sota{58.9$_\text{(10.2)}$}\\
\bottomrule
\end{tabular}}
\caption{\small Performance comparison on \textbf{general-domain QA datasets}. Datasets marked with \textsuperscript{\textdagger} are the source of training data used by Search-R1 and \texttt{s3}. We show generation accuracy (\S\ref{subsec:metric}) as the main results, and exact match scores in brackets. We use E5-base-v2 as the retriever and Wikipedia-2018 as the corpus. ``Searcher'' shows the number of parameters of the searcher model. ``\#Train'' shows the amount of training data used to train the searcher. DeepRetrieval$_\text{BM25}$ is trained on NQ, Search-R1 and \texttt{s3} are trained on NQ+HotpotQA with different training size (170k vs 2.4k). Results are averaged by three runs.}
\vspace{-1em}
\label{tab:perf_general}
\end{table*}

\begin{table*}[t]
\vspace{-2em}
\centering
\small
\resizebox{\textwidth}{!}{
\begin{tabular}{lrrcccccc}
\toprule
&&& \multicolumn{5}{c}{\textbf{Medical RAG-QA Datasets (MIRAGE)}}\\
\cmidrule(lr){4-8}
\textbf{Methods} &\textbf{Searcher} &\textbf{\#Train} & \textbf{MedQA-US} & \textbf{MedMCQA} & \textbf{PubMedQA} & \textbf{BioASQ-Y/N} &\textbf{MMLU-Med} &  \textbf{Avg.} \\
\midrule
\#Test Data && &1,273 &4,183 &500 &618 &1,089\\
\midrule
w/o retrieval &- &0  &61.7$_\text{(45.8)}$ &55.8$_\text{(29.3)}$ &55.6$_\text{(0.0)}$ &76.9$_\text{(0.0)}$ &76.4$_\text{(35.8)}$ &65.3$_\text{(22.2)}$\\
\midrule
\rowcolor{orange!10}
\multicolumn{9}{c}{\textbf{{Corpus: Wikipedia 2018}}~\cite{karpukhin2020dense}} \\
RAG$_\text{BM25}$  &- &0  &61.6$_\text{(48.2)}$ &57.5$_\text{(45.2)}$ &52.8$_\text{(4.6)}$ &73.6$_\text{(6.3)}$ &77.6$_\text{(61.9)}$ &64.6$_\text{(33.2)}$\\
DeepRetrieval$_\text{BM25}$ &3B &70k &62.5$_\text{(45.4)}$ &61.3$_\text{(44.8)}$ &56.2$_\text{(8.2)}$ &\sota{77.3$_\text{(9.2)}$} &\sota{79.2$_\text{(57.9)}$} &67.3$_\text{(33.1)}$\\
RAG$_\text{E5}$  &- &0    &61.5$_\text{(46.7)}$  &58.0$_\text{(44.7)}$  &54.6$_\text{(3.8)}$  &73.3$_\text{(5.3)}$ &77.9$_\text{(62.2)}$ &65.1$_\text{(32.5)}$\\
IRCoT &7B &0  &62.8$_\text{(45.1)}$ &60.5$_\text{(45.4)}$ &54.2$_\text{(8.6)}$ &73.0$_\text{(13.8)}$ &78.7$_\text{(58.2)}$ &65.8$_\text{(34.2)}$\\
IRCoT  &14B &0  &61.7$_\text{(48.9)}$ &60.3$_\text{(46.7)}$ &53.0$_\text{(7.6)}$ &75.2$_\text{(11.8)}$ &77.2$_\text{(61.9)}$ &65.5$_\text{(35.4)}$\\
Search-o1  &14B & 0 &64.5$_\text{(55.4)}$ &59.6$_\text{(47.7)}$ &52.2$_\text{(1.8)}$ &74.9$_\text{(0.2)}$ &77.7$_\text{(63.9)}$ &65.8$_\text{(33.8)}$\\
Search-R1-3B (Ret)  &3B &170k  &58.8$_\text{(47.2)}$ &53.7$_\text{(41.4)}$ &53.8$_\text{(4.4)}$ &63.6$_\text{(4.4)}$ &68.4$_\text{(55.4)}$ &59.7$_\text{(30.6)}$\\
Search-R1-7B (Ret)  &7B &170k  &62.6$_\text{(45.7)}$ &59.2$_\text{(42.8)}$ &55.4$_\text{(5.2)}$  &71.2$_\text{(6.5)}$ &69.3$_\text{(53.3)}$ &63.5$_\text{(30.7)}$\\
\hdashline[2pt/2pt] \\ [-8pt]
\texttt{s3} &7B &2.4k &\sota{65.7$_\text{(47.1)}$} &\sota{61.5$_\text{(44.3)}$} &\sota{56.6$_\text{(5.2)}$} &\sota{77.3$_\text{(7.1)}$} &76.0$_\text{(56.3)}$ &\sota{68.3$_\text{(32.0)}$}\\
\midrule
\rowcolor{green!10}
\multicolumn{9}{c}{\textbf{Corpus: Wikipedia+PubMed+Textbook}~\cite{xiong2024benchmarking}} \\
RAG$_\text{BM25}$ &- &0  &65.4$_\text{(43.1)}$ &59.9$_\text{(44.4)}$ &79.4$_\text{(10.8)}$ &88.4$_\text{(6.5)}$ &79.6$_\text{(57.1)}$ &74.5$_\text{(32.4)}$\\
DeepRetrieval$_\text{BM25}$ &3B &70k &65.0$_\text{(35.1)}$ & 65.1$_\text{(44.2)}$ &78.6$_\text{(16.2)}$ &89.5$_\text{(7.4)}$ &79.3$_\text{(49.1)}$ &75.8$_\text{(30.4)}$\\
RAG$_\text{E5}$  &- &0   &64.1$_\text{(43.4)}$  &60.1$_\text{(45.0)}$ &79.4$_\text{(10.8)}$ &89.8$_\text{(5.0)}$ &78.8$_\text{(58.8)}$ &74.6$_\text{(32.6)}$\\
IRCoT &7B &0  &63.9$_\text{(38.6)}$ &62.7$_\text{(45.3)}$ &75.4$_\text{(13.0)}$ &87.2$_\text{(5.8)}$ &\sota{79.7$_\text{(54.9)}$} &73.8$_\text{(31.5)}$\\
IRCoT  &14B &0  &62.7$_\text{(43.8)}$  &62.3$_\text{(46.6)}$ &74.0$_\text{(10.8)}$ &87.9$_\text{(5.3)}$ &79.6$_\text{(59.0)}$ &73.3$_\text{(33.1)}$\\
Search-o1  &14B & 0 &65.0$_\text{(50.1)}$ &61.1$_\text{(47.6)}$ &74.2$_\text{(12.0)}$ &89.3$_\text{(5.3)}$ &78.1$_\text{(59.5)}$ &73.5$_\text{(34.1)}$\\
Search-R1-3B (Ret)  &3B &170k  &57.5$_\text{(45.5)}$ &54.8$_\text{(40.7)}$ &71.4$_\text{(7.8)}$ &73.3$_\text{(3.6)}$ &62.0$_\text{(47.6)}$ &63.8$_\text{(29.0)}$\\
Search-R1-7B (Ret)  &7B &170k  &62.1$_\text{(43.2)}$ &61.9$_\text{(44.2)}$ &78.6$_\text{(8.0)}$ &86.3$_\text{(5.3)}$ &69.9$_\text{(48.9)}$ &71.8$_\text{(29.9)}$\\
\hdashline[2pt/2pt] \\ [-8pt]
\texttt{s3} &7B &2.4k &\sota{65.7$_\text{(45.7)}$} &\sota{65.3$_\text{(45.4)}$} &\sota{81.5$_\text{(13.6)}$} &\sota{92.1$_\text{(6.5)}$} &78.3$_\text{(56.2)}$ &\sota{76.6$_\text{(33.5)}$}\\
\bottomrule
\end{tabular}}
\caption{\small \textbf{Performance on medical-domain QA datasets}~\cite{xiong2024benchmarking}, using \textit{Claude-3-Haiku} as the generator. We report \emph{judge\_check} as the primary metric (see \S\ref{subsec:metric}), with exact match in brackets. Retrieval is performed with E5-base-v2 under two corpus settings: \textbf{Wikipedia-2018} and \textbf{Wikipedia+PubMed+Textbook}. \texttt{s3} achieves the highest overall accuracy among all retrieval-augmented methods in both settings. None of the methods is trained on medical data: DeepRetrieval$_\text{BM25}$ is trained on 70k NQ, Search-R1 on 170k NQ+HotpotQA, and \texttt{s3} on 2.4k NQ+HotpotQA. Results are averaged by three runs.}

\label{tab:perf_medical}
\end{table*}

\section{Experiments}
\subsection{Experimental Setups}
\noindent\textbf{Evaluation Metric.}
\label{subsec:metric}
We measure performance using \emph{Generation Accuracy}, which combines a fast span-matching test~\cite{ma2021replication,lin2021pyserini} with a lightweight LLM-based correctness check (Figure~\ref{fig:answer_check}). Given a model prediction $p$ and a set of gold answers $\mathcal{A}$, we compute:
\begin{align}
\label{eq:gen_acc}
    \text{GenAcc} = \text{span\_check} \lor \text{judge\_check}
\end{align}
which can be either 1 or 0, determined by the following evaluation flow:

\vspace{-0.5em}
\begin{tcolorbox}[colback=blue!2!white, colframe=blue!50!black, title= \textbf{Evaluation Flow of Generation Accuracy}]
\small
\textbf{Input:} Prediction $p$, Gold Answers $\mathcal{A}$

\vspace{4pt}
\noindent
\textbf{Step 1:} Normalize $p$ and $\mathcal{A}$ (lowercase, remove punctuation and articles). \\
\textbf{Step 2:} \texttt{span\_check} $\rightarrow$ If any $a \in \mathcal{A}$ is a token span in $p$, return \texttt{GenAcc} = 1. \\
\textbf{Step 3:} \texttt{judge\_check} $\rightarrow$ Prompt LLM: \textit{``Does $p$ contain any of $\mathcal{A}$?''} \\
\textbf{Step 4:} Return \texttt{GenAcc} = 1 if LLM says yes; else 0.
\end{tcolorbox}

\vspace{-1em}
\begin{tcolorbox}[colback=gray!10,colframe=black!50,title= Why Exact Match Falls Short - An Example]
\footnotesize

\textbf{Golden answer:} \texttt{"Barack Obama"}\\
\textbf{LLM response:} 
\texttt{"The 44th President of the United States was Barack Obama."}\\
\textbf{Exact match:} 0 \quad (response $\neq$ golden)\\
\textbf{Generation Accuracy:} 1 \quad (span\_check succeeds)
\end{tcolorbox}
\noindent We choose this metric because it better captures semantic correctness and aligns more closely with human judgment than traditional exact match (see Appendix~\ref{ap:alignment} for supporting evidence).

\noindent\textbf{Datasets.} Following prior study~\cite{jin2025search}, we construct the training set by combining samples from Natural Questions (NQ) and HotpotQA. Since \texttt{span\_check} may incorrectly accept answers for questions with semantic negation (e.g., treating “not true” as matching “true”), we remove all yes/no and true/false questions from the training set to ensure reliable reward signals. We evaluate such questions in test data w/ LLMJudge.
To focus training on harder examples, we filter out samples where the generator LLM (Qwen2.5-14B-Instruct) already produces a correct answer using naïve RAG retrieval. This reduces the dataset size from 169,615 to 70,286. As later shown in Figure~\ref{fig:reward_curves}, \texttt{s3} rapidly converges within $\sim$15 training steps. 
For evaluation, we use the checkpoints at step 20. Given a batch size of 120, this corresponds to approximately 2.4k training examples, highlighting the data efficiency of our method. We evaluate on six general-domain QA benchmarks: NQ~\cite{kwiatkowski2019natural}, TriviaQA~\cite{joshi2017triviaqa}, PopQA~\cite{mallen2023llm_memorization}, HotpotQA~\cite{yang2018hotpotqa}, 2WikiMultihopQA~\cite{xanh2020_2wikimultihop}, and Musique~\cite{trivedi2021musique}, as well as MIRAGE~\cite{xiong2024benchmarking}, a suite of five medical-domain QA datasets.

\begin{table*}[h]
\vspace{-1em}
\centering
\small
\resizebox{\textwidth}{!}{
\begin{tabular}{ccccccccccl}
\toprule
&&& \multicolumn{3}{c}{\textbf{Single-Hop}} &\multicolumn{3}{c}{\textbf{Multi-Hop}} \\
\cmidrule(lr){4-6} \cmidrule(lr){7-9} 
\#Retrieval $\rightarrow$ \#Select &\#Turns &\#MaxContexts & \textbf{NQ\textsuperscript{\textdagger}} & \textbf{TriviaQA} & \textbf{PopQA} & \textbf{HotpotQA\textsuperscript{\textdagger}} & \textbf{2wiki} & \textbf{Musique} & \textbf{Avg.} \\
\midrule
8 $\rightarrow$ 3 &3 &9 &{70.5$_\text{(3.2)}$} &{84.0$_\text{(24.6)}$} &{57.7$_\text{(5.9)}$} &{62.4$_\text{(11.1)}$} &{52.4$_\text{(8.3)}$} &26.2$_\text{(7.9)}$ &58.9$_\text{(10.2)}$\\
5 $\rightarrow$ 3 &3 &9 &69.6$_\text{(3.5)}$ &83.4$_\text{(24.3)}$ &57.4$_\text{(5.8)}$ &62.0$_\text{(11.9)}$ &53.8$_\text{(7.8)}$ &24.5$_\text{(2.3)}$ &58.5$_\text{(9.3)}$\\
5 $\rightarrow$ 3 &4 &12 &70.0$_\text{(3.5)}$ &83.8$_\text{(24.8)}$ &57.7$_\text{(5.8)}$ &62.5$_\text{(12.3)}$ &54.7$_\text{(8.0)}$ &25.7$_\text{(3.2)}$ &59.1$_\text{(9.6)}$\\
3 $\rightarrow$ 3 &4 &12 &68.9$_\text{(3.7)}$ &82.0$_\text{(24.9)}$ &56.4$_\text{(6.1)}$ &62.0$_\text{(11.9)}$ &51.7$_\text{(7.7)}$ &24.7$_\text{(2.8)}$ &57.7$_\text{(9.5)}$\\
3 $\rightarrow$ 3 &3 &9  &69.4$_\text{(3.5)}$ &82.3$_\text{(24.4)}$ &57.0$_\text{(5.7)}$ &61.8$_\text{(11.7)}$ &51.5$_\text{(8.2)}$ &25.1$_\text{(2.3)}$ &57.9$_\text{(9.3)}$\\

\bottomrule
\end{tabular}}
\caption{Study of the numbers of retrieved documents (\#Retrieval) and turns (\#Turns). Maximum selection is set to 3 across all settings. We use the frozen Claude-3-Haiku as the generator LLM for this study.}
\label{tab:doc_turn}
\vspace{-1em}
\end{table*}

\noindent\textbf{Baselines.}
We compare \texttt{s3} against diverse RAG systems:
\noindent\textbf{(1) End-to-End Fine-tuning.}
Fully fine-tuned models that jointly retrieve and generate using outcome-based RL or supervision: {Search-R1 (3B/7B)}, {SFT (3B)}, and {R1 (7B)} where 3B/7B for SFT and R1 are based on Qwen2.5-3B/7B-Instruct. 
\textbf{(2) Static Retrieval+Frozen Generator.}
Methods that retrieve documents using a fixed or scripted strategy, then pass them to a frozen generator:
{RAG-BM25}, {RAG-E5}: retrieval via BM25 or E5-base~\cite{wang2022text}.
{DeepRetrieval-BM25 (3B)}: RL-trained searcher optimizing recall, paired with BM25.
\textbf{(3) Active Retrieval+Frozen Generator.}
A diagnostic setting where we extract the documents retrieved during a model’s reasoning trajectory and feed them to a frozen generator:
{Search-R1-3B/7B (Ret)}, {IRCoT}, and Search-o1 all fall under this category, differing only in whether retrieval is learned (Search-R1) or prompted (IRCoT~\cite{trivedi-etal-2023-interleaving}, Search-o1~\cite{li2025search}).
See more details in Appendix~\ref{apsub:baselines}.

\noindent\textbf{Models for Training and Evaluation.}
Throughout all the training processes, we use Qwen2.5-7B-Instruct~\cite{yang2024qwen2} as the base searcher LLM to train, and use
Qwen2.5-14B-Instruct\footnote{In this paper, we use ``GPTQ-Int4'' version of ``Qwen2.5-14B-Instruct'' for its high efficiency. We deploy frozen LLMs using vLLM~\cite{kwon2023efficient} for fast inference.} as the frozen generator for both answer generation and \texttt{judge\_check} for reward computation. For evaluation, we use Claude-3-Haiku as the LLM for  \texttt{judge\_check} to ensure high evaluation quality. We test three frozen generators: Qwen2.5-7b/14b-Instruct and Claude-3-Haiku.
\
Both training and evaluation are conducted on five NVIDIA A100 80GB PCIe GPUs. RAGEN~\cite{wang2025ragen} and VERL~\cite{sheng2024hybridflow} are used as the base architecture for multi-turn RL training. We place more details in Appendix~\ref{apsub:setup}. 

\section{Results}

We evaluate \texttt{s3} across six general-domain and five medical-domain QA benchmarks, with frozen generators ranging from Qwen2.5-7B/14B to Claude-3-Haiku. We report generation accuracy as the primary metric and provide detailed comparisons across baselines, reward functions, and training efficiency.

\noindent{}\textbf{General Domain RAG Performance.}
Table~\ref{tab:perf_general} summarizes results across general QA datasets. \texttt{s3} achieves the highest average accuracy of 58.9\%, outperforming all static, zero-shot, and end-to-end tuned baselines. This is particularly notable given its extreme data efficiency—trained on just 2.4k examples, compared to 70k for DeepRetrieval and 170k for Search-R1.

\begin{tcolorbox}[colback=orange!4, colframe=purple!170, title= \textbf{Takeaway \#1: Searcher-Only is better than End-to-End Optimization for RAG}]
\texttt{s3} consistently outperforms Search-R1 on search quality, revealing that most of the performance gain in RAG stems from improving the search capability instead of aligning generation outputs.
\end{tcolorbox}

\noindent Compared to IRCoT-14B, which conducts zero-shot retrieval with 2× the parameter count, \texttt{s3} gains +4.6 points on average. Relative to Search-R1-7B (Ret), which uses the same backbone, \texttt{s3} improves by +1.5 points while avoiding any generator tuning. These gains are consistent across both single-hop (e.g., 70.0\% on NQ) and multi-hop datasets (e.g., 62.4\% on HotpotQA), showing that learned search behavior transfers across reasoning complexity.

\noindent\textbf{Medical Domain QA Performance.}
Table~\ref{tab:perf_medical} reports performance on the MIRAGE suite~\cite{xiong2024benchmarking} under both corpus settings. \texttt{s3} achieves the highest average accuracy (76.6\%) when using the combined Wikipedia+PubMed+Textbook corpus, surpassing all retrieval-augmented baselines.

Interestingly, while Search-R1 shows competitive scores on Wikipedia-only corpora, its performance deteriorates on richer corpora, indicating overfitting to shallow heuristics or memorized formats. In contrast, \texttt{s3} and DeepRetrieval remain robust, with \texttt{s3} achieving 81.5\% on PubMedQA and outperforming IRCoT across four of five tasks.

\begin{tcolorbox}[colback=orange!4, colframe=purple!170, title=\textbf{Takeaway \#2: Searcher-Only Training enables Domain Transfer}]
\texttt{s3}'s zero-shot success on medical QA, despite training only on general QA, suggests that reinforcement-learned search skills generalize more reliably than generation-tuned approaches.
\end{tcolorbox}

\begin{figure}[t]
\centering
\includegraphics[width=\linewidth]{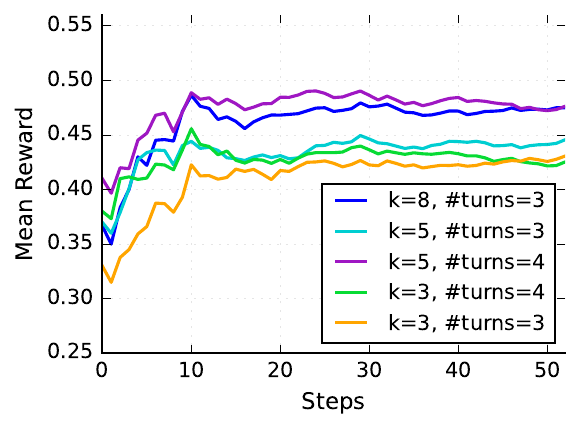}
\vspace{-2em}
\caption{Reward Curves for top $k=\{3,5,8\}$ and $\#\text{turns}=\{3,4\}$. The maximum selection is kept as 3.}
\label{fig:reward_curves}
\vspace{-1em}
\end{figure}
\begin{figure*}[t!]
\vspace{-1.5em}
    \centering
    \includegraphics[width=\linewidth]{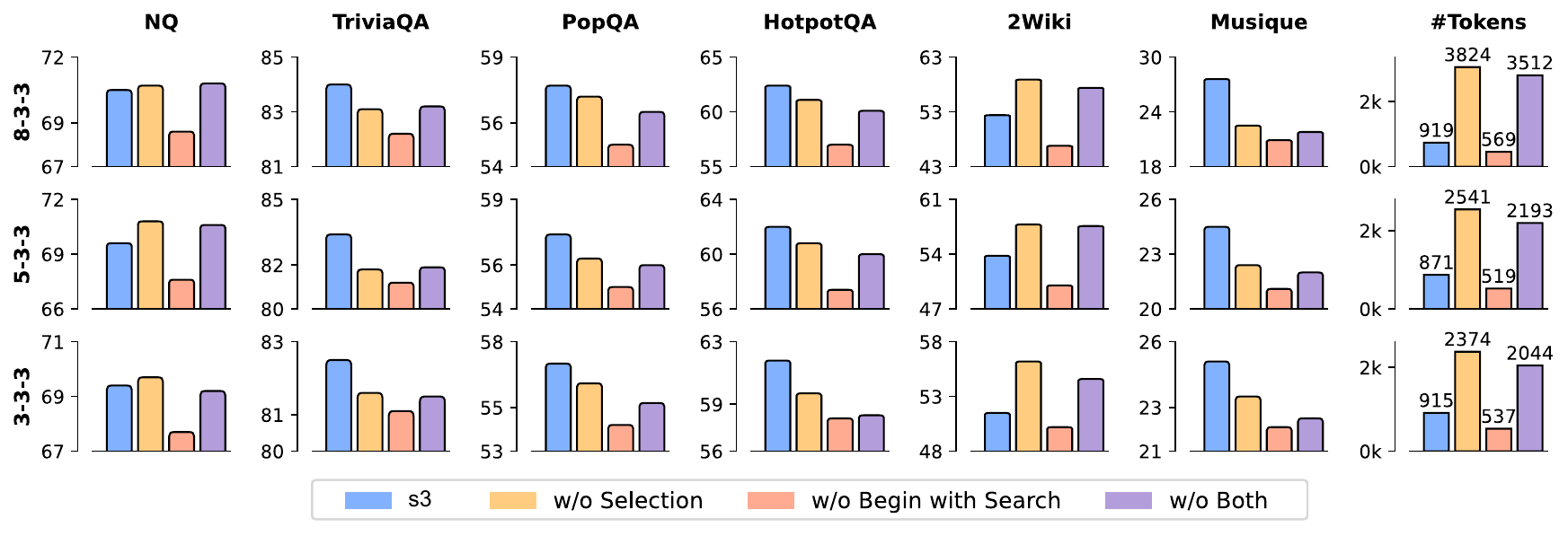}
    \vspace{-2em}
\caption{\small
\textbf{Ablation study on s3 components.} Each row corresponds to a different configuration of Retrieval:Selection:Turns = 8:3:3, 5:3:3, and 3:3:3. The first six columns report generation accuracy. \textbf{“Begin with Search”} refers to initializing the first query with the original question. \textbf{“Document Selection”} refers to the selection step within the s3 loop (Step 3). We observe that removing Begin with Search leads to a significant drop in performance. While removing Document Selection sometimes yields better performance, the full s3 system still performs competitively—and most importantly, drastically reduces input token usage (2.6$\times$ $\sim$ 4.2$\times$ less tokens), improving overall efficiency.}
    \label{fig:ablation}
    \vspace{-1em}
\end{figure*}
\noindent\textbf{Retrieval Behavior and Search Dynamics}
We analyze the effect of retrieval parameters (\#retrieved documents and \#turns) in Table~\ref{tab:doc_turn} and reward progression in Figure~\ref{fig:reward_curves}. \texttt{s3} reaches peak performance with ($k$=8, turns=3), and adding more turns or broader retrieval brings limited improvement. This indicates that the policy rapidly learns to emit focused and early queries, capturing most useful content without unnecessary expansion. 

\noindent\textbf{Training Efficiency}
Table~\ref{tab:efficiency_study} shows that it takes 20 PPO steps (2.4k examples) to train \texttt{s3}, while Search-R1 requires 2,100 steps (170k examples). Even accounting for the higher per-step cost due to LLM-based reward computation, the total wall-clock time is reduced by $\sim$33×. Moreover, \texttt{s3} avoids retriever pretraining and operates with a smaller 7B policy model, making it a practical method for low-resource RL training.
\texttt{s3} achieves state-of-the-art performance with orders of magnitude less data and compute, suggesting a more sustainable path for RAG optimization.
\begin{table}[t]
\centering
\small
\resizebox{\linewidth}{!}{
\begin{tabular}{lccc}
\toprule
&\textbf{Time/Step} &\textbf{Training Steps} &\textbf{Total}  \\
\midrule
Search-R1 &1.8m &$\sim$2,100 &3,780m\\
DeepRetrieval$_\text{BM25}$ &1.3m &$\sim$1,600 &2,080m\\
\texttt{s3} &5.7m &$\sim$20 &\textcolor{purple}{\textbf{114m}}\\

\bottomrule
\end{tabular}}
\caption{\small Comparison of Training Efficiency (tested with batch size=120 on five NVIDIA A100 GPUs). Note: \texttt{s3} is slower stepwise since we need to conduct generation and evaluation by a frozen LLM for reward computation during training.}
\label{tab:efficiency_study}
\vspace{-0.5em}
\end{table}

\noindent\textbf{Reward Function Comparison}
Table~\ref{tab:reward_study} compares different reward signals used for computing GBR. LLMJudge provides slightly higher final scores, but is too costly for scalable training. In contrast, GenAcc offers strong performance while remaining efficient and aligning better with human evaluation than EM or span-based heuristics. Appendix~\ref{ap:alignment} shows that GenAcc matches human judgment on 96.4\% of samples, while Exact Match used by Search-R1 captures only 15.8\%.
\vspace{-0.3em}
\begin{tcolorbox}[colback=orange!4, colframe=purple!170, title=\textbf{Takeaway \#3: Reward Choice directly shapes Search Quality}]
Using semantically or human preference aligned metrics like our GenAcc (\S\ref{subsec:metric}) encourages the search policy to retrieve substantively helpful documents, rather than optimizing for brittle string overlap.
\end{tcolorbox}
\begin{table}[h]
\centering
\small
\resizebox{\linewidth}{!}{
\begin{tabular}{lcccc}
\toprule
&\textbf{GenAcc} &\textbf{LLMJudge} &\textbf{Span} & \textbf{EM}  \\
\midrule
General QA &58.9 &59.6 &57.1 &50.5 \\
Medical QA &76.6 &77.3 &74.3 &70.3 \\

\bottomrule
\end{tabular}}
\caption{\small Comparison of RAG performance under different reward functions. \textit{LLMJudge} (judge\_check) yields the highest scores but is computationally expensive. \textit{GenAcc} offers a good balance of accuracy and efficiency, while \textit{Span} (span\_check) and \textit{EM} underperform due to limited semantic coverage.}
\vspace{-1em}

\label{tab:reward_study}
\end{table}


\noindent\textbf{Effects of Selection and ``Begin with Search''.} We investigate the role of two components in the \texttt{s3} loop: document selection and initialization with the original question (Begin with Search''). As shown in Figure~\ref{fig:ablation}, removing the selection step degrades performance on four out of six datasets. This is expected, as passing all retrieved documents to the generator increases token length, up to 4$\times$ with $k=8$, and introduces more noise. Still, performance improves slightly on NQ and 2Wiki, likely because broader context benefits multi-hop reasoning or compensates for overly aggressive pruning.
Disabling ``Begin with Search'' consistently causes a significant drop, underscoring the importance of seeding the search process with a strong initial query. Interestingly, when both selection and initialization are removed, performance recovers slightly compared to removing only initialization. This suggests that selection and initialization interact conditionally—selection may amplify the downsides of poor initialization by prematurely filtering out useful context.

\section{Conclusion}
We present \texttt{s3}, a framework that trains a search agent using the Gain Beyond RAG reward. By decoupling search from generation and optimizing only the retriever, \texttt{s3} outperforms strong baselines with just 2.4k examples. Our results show that targeted search policy learning yields substantial gains in both efficiency and generalization, offering a scalable path for improving RAG systems.

\section{Limitations}

While \texttt{s3} demonstrates strong empirical performance with remarkable data efficiency, several limitations warrant discussion.

\paragraph{Dependency on Frozen Generators.} Our framework assumes the availability of a capable frozen generator LLM. Although this enables model-agnostic training, it implicitly relies on the generator's ability to make use of improved context. For lower-capacity or instruction-weak generators, the gains from better retrieval may not fully translate into better outputs.

\paragraph{Reward Estimation Bottleneck.} The use of generation-based rewards such as GenAcc necessitates LLM inference during training to compute reward signals. This introduces computational overhead compared to token-level or retrieval-only objectives, limiting scalability. Although we show that \texttt{s3} achieves high performance with minimal steps, online reward computation remains more costly than offline retrieval optimization.

\paragraph{Broader Impacts.}
On the positive side, \texttt{s3} reduces the data and compute burden for training effective retrieval agents, making RAG systems more accessible to low-resource communities. It may also benefit domains such as healthcare or scientific QA where labeled data is scarce. However, like all retrieval-augmented systems, \texttt{s3} inherits the biases of both its searcher and generator. If deployed without careful curation of source corpora, it may propagate misinformation or reflect existing societal biases. We encourage practitioners to audit both retrieval sources and downstream outputs when applying this framework in sensitive domains.

Overall, while \texttt{s3} advances the state of search-agent training, further work is needed to address these limitations and ensure safe, robust deployment in real-world settings.

\section*{Acknowledgements}
Research was supported in part by National Science Foundation IIS-19-56151, NSF IIS 25-37827, the Molecule Maker Lab Institute: An AI Research Institutes program supported by NSF under Award No. 2019897, and the Institute for Geospatial Understanding through an Integrative Discovery Environment (I-GUIDE) by NSF under BRIES Program No. HR0011-24-3-0325.

\bibliography{references}

\clearpage
\appendix
\section*{Contents of Appendix}
\noindent A. Implementation Details \dotfill \pageref{ap:impl_details}\\
\quad A.1 Baselines Details \dotfill \pageref{apsub:baselines}\\
\quad A.2 Setup Details \dotfill \pageref{apsub:setup}\\
\quad A.3 Datasets \& Corpora \dotfill
\pageref{subsec:datasets}\\
\quad A.4 Generation Accuracy Computation \dotfill \pageref{subsec:genacc_compute}\\
\quad A.5 Document Extraction Logic \dotfill \pageref{subsec:doc_extraction}\\
B. Alignment Study of Evaluation Metrics \dotfill \pageref{ap:alignment}\\
C. Prompts \dotfill \pageref{ap:prompts}\\
D. Scalability Study \dotfill \pageref{sec:scalability}\\

\section{Implementation Details}
\label{ap:impl_details}
For static retrieval baselines running on MIRAGE, we use the question itself instead of question+options to retrieve.

\subsection{Baselines Details}
\label{apsub:baselines}

\textbf{IRCoT (7B and 14B).} IRCoT\footnote{\url{https://github.com/StonyBrookNLP/ircot}}~\cite{trivedi-etal-2023-interleaving} is a prompting-based method that alternates between chain-of-thought reasoning and retrieval. It requires no fine-tuning: the model is instructed via prompt to iteratively reason about a question and issue retrieval queries, integrating newly retrieved evidence into its reasoning process. We apply IRCoT using both Qwen2.5-7B-Instruct and Qwen2.5-14B-Instruct.

\noindent\textbf{DeepRetrieval-BM25-3B~\cite{jiang2025deepretrieval}.} This baseline employs a 3B-parameter language model trained with reinforcement learning on retrieval metrics such as recall and NDCG. It learns to generate search queries that maximize the retrieval of relevant documents using a BM25 search engine. Training is conducted on 70k QA examples in NQ dataset with answer span reward (evidence-seeking task in \citep{jiang2025deepretrieval}), focusing exclusively on improving retrieval performance, not generation. We use its publicly released checkpoint\footnote{\url{https://huggingface.co/DeepRetrieval/DeepRetrieval-NQ-BM25-3B}}.

\noindent\textbf{Search-R1-3B and Search-R1-7B ~\cite{jin2025search}.} These baselines\footnote{\url{https://huggingface.co/PeterJinGo/SearchR1-nq_hotpotqa_train-qwen2.5-3b-em-ppo}, \url{https://huggingface.co/PeterJinGo/SearchR1-nq_hotpotqa_train-qwen2.5-7b-em-ppo}} use 3B and 7B parameter models, respectively, and are trained end-to-end to jointly retrieve and generate answers. Reinforcement learning is applied on 170k training examples, using an exact match (EM) reward to guide both retrieval query formulation and answer generation. The model directly integrates search results into its reasoning steps within a single retrieval round.

\noindent\textbf{Search-o1.} Search-o1~\cite{li2025search} is an inference-time retrieval controller designed to enhance long-form reasoning in o1-style models such as QwQ and OpenAI’s o1-preview. It is not trained with reinforcement learning or fine-tuned at all. Instead, Search-o1 leverages frozen LLMs and augments them with retrieval by prompting the model to emit search queries mid-reasoning, enclosed in special tokens (e.g., \texttt{<|begin\_search\_query|>}... Retrieved documents are then post-processed using a Reason-in-Documents module before being injected back into the reasoning flow.

\noindent\textbf{RAG-BM25 and RAG-E5~\cite{lewis2020retrieval}.} These are naive retrieval-augmented generation baselines with no model training. RAG-BM25 uses top-$k$ documents retrieved from a BM25 index, while RAG-E5 retrieves passages using dense retrieval based on E5 embeddings. In both settings, the retrieved documents are prepended to the input prompt and fed into a frozen generator LLM. We set $k=3$, following prior study~\cite{lin2023ra,jin2025search}.

\noindent\textbf{SFT and R1.} On general-domain RAG datasets, we train an SFT model with Qwen2.5-3B-Instruct using the same dataset as Search-R1's 170k NQ+HotpotQA with TRL~\cite{vonwerra2022trl} framework. R1 is the ``no search'' version of Search-R1~\cite{jin2025search}, replicating Deepseek-R1-Zero~\cite{guo2025deepseek} with a small LLM. We use its publicly released checkpoint\footnote{\url{https://huggingface.co/PeterJinGo/R1-nq_hotpotqa_train-qwen2.5-7b-em-ppo-v0.2}}. 

\noindent\textbf{CoT~\cite{wei2022chain} and Direct Inference.} CoT (Chain-of-Thought) prompting instructs the LLM to generate intermediate reasoning steps before producing an answer, without any external retrieval. Direct Inference simply feeds the raw question into the LLM. Neither baseline involves any form of training or finetuning.

To ensure a fair comparison, we set the maximum number of turns to 4 and limit the context to 3 documents per turn for all multi-turn baselines (IRCoT, Search-R1, and Search-o1) and \texttt{s3}, aligning with prior study~\cite{jin2025search}.

\subsection{Setup Details}
\label{apsub:setup}
\noindent\textbf{Hardware.} All training and evaluation processes are run on five NVIDIA A100 80GB PCIe on a system with an AMD EPYC 7513 32-Core Processor and 1.0 TB of RAM.

\noindent\textbf{Software.}
We built \texttt{s3} using Python 3.9, leveraging the VERL framework~\citep{sheng2024hybridflow}\footnote{\url{https://github.com/volcengine/verl}} (v0.1) as the backbone for reinforcement learning with language models, and RAGEN~\cite{wang2025ragen}\footnote{\url{https://github.com/RAGEN-AI/RAGEN}} as the underlying multi-turn RL architecture. Our implementation uses vLLM (v0.8.5)~\citep{kwon2023efficient} for fast LLM inference and evaluation, PyTorch (v2.4.0) with CUDA 12.1 for deep learning, and Ray~\citep{moritz2018ray} for distributed training and serving. To improve performance, we integrate Flash Attention 2~\citep{dao2023flashattention} for efficient attention computation, PySerini (v0.22.1)~\citep{lin2021pyserini} for retrieval and evaluation, and FAISS-GPU (v1.7.2)~\citep{douze2024faiss} for high-speed dense retrieval.

\noindent\textbf{Model parameters.}  
We fine-tune Qwen2.5-7B-Instruct using Proximal Policy Optimization (PPO) via VERL. Training is conducted with a total batch size of 120, using micro-batches of size 15 for the actor and 10 for the critic, and a rollout temperature of 0.6. The actor and critic learning rates are set to $1 \times 10^{-6}$ and $1 \times 10^{-5}$, respectively, with no warm-up for the actor and a 1\% warm-up ratio for the critic. Both models use gradient checkpointing and parameter offloading to reduce memory overhead. Following prior work~\cite{jin2025search}, we adopt XFORMERS~\cite{xFormers2022} as the attention backend in vLLM and enable state masking to prevent incorrect supervision signals. KL regularization is applied with a coefficient of 0.001. 
For answer generation and LLM-based \texttt{judge\_check} during training, we run Qwen2.5-14B-Instruct-GPTQ-Int4\footnote{\url{https://huggingface.co/Qwen/Qwen2.5-14B-Instruct-GPTQ-Int4}} on a dedicated A100 80GB GPU with vLLM. The retriever (E5-base) is deployed alongside PySerini on the same five GPUs used for PPO training. The context window is set to 8,000 tokens, with a maximum of 1,400 tokens allocated to the top-$k$ retrieved documents per turn. 

\subsection{Datasets \& Corpora}
\label{subsec:datasets}

\noindent\textbf{Datasets.}  
We evaluate on six general-domain QA datasets and five medical-domain QA datasets. 

\noindent\textbf{\textit{General-domain}} datasets include {Natural Questions (NQ)}~\cite{kwiatkowski2019natural}, {TriviaQA}~\cite{joshi2017triviaqa}, {PopQA}~\cite{mallen2023llm_memorization}, {HotpotQA}~\cite{yang2018hotpotqa}, {2WikiMultihopQA}~\cite{xanh2020_2wikimultihop}, and {Musique}~\cite{trivedi2021musique}. \footnote{All the general-domain QA datasets are available at \url{https://huggingface.co/datasets/RUC-NLPIR/FlashRAG_datasets}.}

\noindent For \textbf{\textit{medical-domain}}, we adopt the MIRAGE benchmark~\cite{xiong2024benchmarking}, which includes five datasets: {MedQA-US}~\cite{jin2021disease}, {MedMCQA}~\cite{pal2022medmcqa}, {PubMedQA*}~\cite{jin2019pubmedqa}, {BioASQ-Y/N}~\cite{tsatsaronis2015overview, krithara2023bioasq}, and {MMLU-Med}~\cite{hendrycks2020measuring}.\footnote{All the medical-domain QA datasets are available at \url{https://github.com/Teddy-XiongGZ/MIRAGE/blob/main/benchmark.json}.}

\paragraph{Corpora.}  
\textbf{\textit{For general-domain QA}}, we follow prior work~\cite{jin2025search} and use the {Wikipedia 2018 dump}~\cite{karpukhin2020dense} as the sole knowledge source.\footnote{Wikipedia 2018 dump is available at \url{https://huggingface.co/datasets/RUC-NLPIR/FlashRAG_datasets} (retrieval-corpus folder)}  
\textbf{\textit{For medical-domain QA}}, we evaluate under two corpus settings: (1) the {Wikipedia 2018} dump~\cite{karpukhin2020dense} alone, and (2) a composite biomedical corpus introduced by~\cite{xiong2024benchmarking}, which combines {Wikipedia}, {PubMed}, and {textbook} documents to provide broader domain coverage.\footnote{All the corpora for medical RAG are available at \url{https://huggingface.co/MedRAG}.}

\paragraph{Use of Artifacts.}
All datasets and models are used strictly within research contexts, consistent with their intended use and licensing. Our derived artifacts (e.g., retrieved documents, trained models) are likewise restricted to non-commercial academic use.

\subsection{Generation Accuracy Computation}
\label{subsec:genacc_compute}

To evaluate the effectiveness of retrieval strategies in improving answer generation, we adopt a composite metric called \textbf{Generation Accuracy (GenAcc)}, which is designed to better reflect semantic correctness than surface-form exact match.

\paragraph{Overview.} Given a model prediction $p$ and a set of gold answers $\mathcal{A}$, GenAcc is defined in Eq.~\ref{eq:gen_acc}.
This metric returns $1$ if either a string-normalized token span of any $a \in \mathcal{A}$ is found within $p$, or if a frozen LLM judge deems the answer semantically correct. It returns $0$ otherwise.

\paragraph{1. Span-Based Matching.} 
We first apply a deterministic span check using normalized string comparison. Specifically, we:
\begin{itemize}
  \item Convert both prediction and gold answers to lowercase.
  \item Remove punctuation and articles (\textit{a, an, the}).
  \item Apply whitespace normalization.
\end{itemize}
We then use a tokenizer to compare whether any token span in the prediction matches any normalized gold answer. If a match is found, the score is $1$.

\smallskip
\noindent\textbf{Examples:}
\begin{itemize}[leftmargin=*]
  \item \textit{Success Case:} \\ 
    \textit{\textbf{Prediction:}} \texttt{"The 44th President of the United States was Barack Obama."} \\
    \textit{\textbf{Gold Answer:}} \texttt{"Barack Obama"} \\
    \textit{\textbf{Result:}} Span match succeeds because the normalized gold answer is a token span in the prediction.

  \item \textit{Failure Case (Negation):} \\ 
    \textit{\textbf{Prediction:}} \texttt{"That statement is not true."} \\
    \textit{\textbf{Gold Answer:}} \texttt{"true"} \\
    \textit{\textbf{Result:}} Span match incorrectly succeeds due to token overlap, despite the semantic meaning being opposite. We exclude such yes/no cases from training to avoid this issue.

  \item \textit{Failure Case (Paraphrase):}\\
    \textit{\textbf{Prediction:}} \texttt{"He led the civil rights movement in the 1960s."} \\
    \textit{\textbf{Gold Answer:}} \texttt{"Martin Luther King Jr."} \\
    \textit{\textbf{Result:}} Span match fails because the gold answer does not appear verbatim in the response, even though the answer is implied.
\end{itemize}

\paragraph{2. LLM-Based Semantic Judging.} 
If the span check fails ($0$), we invoke a lightweight correctness check using a frozen LLM (e.g., \texttt{Qwen2.5-14B-Instruct-GPTQ-Int4} for training or \texttt{Claude-3-Haiku} for evaluation). We prompt the model with:
\begin{quote}
\small
\texttt{Please check if any of the golden answers is contained in the following response: \{p\} \\
Golden answers: \{\texttt{str}($\mathcal{A}$)\} \\
Directly answer with 'yes' or 'no'.}
\end{quote}
If the LLM outputs \texttt{yes}, we consider the prediction correct and set the score to $1$.

\smallskip
\noindent\textbf{Examples:}
\begin{itemize}[leftmargin=*]
  \item \textit{Success Case (Numerical Format):} \\
    \textit{\textbf{Prediction:}} \texttt{"The answer is twenty-five."} \\
    \textit{\textbf{Gold Answer:}} \texttt{"25"} \\
    \textit{\textbf{Result:}} Span match fails due to different formats, but the LLM outputs \texttt{yes} based on numerical equivalence.

  \item \textit{Success Case (Units and Symbols):} \\
    \textit{\textbf{Prediction:}} \texttt{"It weighs 3 kilograms."} \\
    \textit{\textbf{Gold Answer:}} \texttt{"3 kg"} \\
    \textit{\textbf{Result:}} Span match fails due to token mismatch, but the LLM recognizes them as equivalent and answers \texttt{yes}.

  \item \textit{Failure Case (Incorrect Entity):} \\
    \textit{\textbf{Prediction:}} \texttt{"The capital of France is Marseille."} \\
    \textit{\textbf{Gold Answer:}} \texttt{"Paris"} \\
    \textit{\textbf{Result:}} Span match fails, and the LLM also outputs \texttt{no}, indicating semantic disagreement.
\end{itemize}

\paragraph{Motivation.} This design avoids brittle behavior from exact match metrics and aligns more closely with human judgments. For instance, if the gold answer is \textit{"Einstein"} and the model prediction is \textit{"Albert Einstein was the scientist who developed the theory of relativity"}, our metric returns 1, while exact match fails due to surface mismatch. Empirically, GenAcc matches human labels on 96.4\% of samples (see Appendix~\ref{ap:alignment}), whereas EM only achieves 15.8\%.

\paragraph{Implementation.} The full reward computing pipeline is implemented through the following components:
\begin{itemize}[leftmargin=*]
  \item \texttt{span\_check}: This function (1) normalizes both prediction and gold answers by applying case-folding, punctuation and article removal, and whitespace normalization; and (2) performs token-level span matching using a tokenizer. This step follows the evaluation strategy introduced in prior work~\cite{ma2021replication} and leverages the \texttt{has\_answer} utility from PySerini\footnote{\url{https://github.com/castorini/pyserini/blob/master/pyserini/eval/evaluate_dpr_retrieval.py}}.
  \item \texttt{judge\_check}: If the span check fails, this fallback invokes a frozen LLM to assess whether the prediction semantically entails any gold answer. The LLM is prompted to respond with a binary judgment ("yes" or "no").
    \item \texttt{check\_answer\_correct}: This function coordinates the evaluation process. It first applies \texttt{span\_check}; if that fails, it falls back to \texttt{judge\_check} for semantic validation. \textbf{Note:} For the medical RAG benchmark (MIRAGE~\cite{xiong2024benchmarking}) evaluation, we exclusively use \texttt{judge\_check}, as most questions are multiple-choice and \texttt{span\_check} can incorrectly accept wrong answers due to its strict matching criteria.
    
\end{itemize}
This hybrid strategy combines the efficiency of lexical matching with the robustness of LLM-based semantic evaluation, ensuring reliable and scalable answer correctness assessment.

\subsection{Document Extraction Logic}
\label{subsec:doc_extraction}
We extract document titles and texts from information blocks using a structured approach that prioritizes important documents. Our extraction algorithm processes text with the following format:
\begin{verbatim}
<information>
Doc 1 (Title: "Document Title 1") ...
Doc 2 (Title: "Document Title 2") ...
</information>
<important_info>
[1, 3]
</important_info>
\end{verbatim}
The algorithm follows these key rules:
\begin{itemize}[leftmargin=15pt]
\item \texttt{<important\_info>} tags apply only to the most recent \texttt{<information>} block
\item If no \texttt{<important\_info>} tag exists for a \texttt{<information>} block, all documents from that block are included
\item Documents are deduplicated based on content
\end{itemize}
The implementation uses regular expressions to:
\begin{enumerate}[leftmargin=*]
\item Identify all information blocks and important document tags
\item Associate each important info tag with its corresponding information block
\item Extract document IDs, titles, and text content
\item Filter documents based on importance markers
\end{enumerate}
The document pattern is matched using a regex that handles variations in spacing and optional quotes around titles.
Our implementation includes appropriate error handling to manage parsing failures and maintains the original order of documents. The algorithm has $O(n)$ time complexity where $n$ is the input string length, with additional factors related to the number of documents and information blocks.

\section{Human Alignment Study of Evaluation Metrics (GenAcc and EM)}
\label{ap:alignment}

To assess the alignment of our primary evaluation metric, \textbf{Generation Accuracy}, with human judgment, we conducted a human annotation study. We randomly sampled 1,000 answer generations from the general-domain QA test set. Each sample was labeled as \texttt{Correct (1)} or \texttt{Incorrect (0)} by human annotators, consisting of two Ph.D. students and one M.S. student majoring in computer science who evenly divided the annotation workload. Figure~\ref{fig:human_eval_instruct} shows the instruction, and the anonymous sheet\footnote{(Anonymous) raw results of human-metric alignment study: \url{https://docs.google.com/spreadsheets/d/e/2PACX-1vQ-aAC6FNJYFJk1Ca8-EGN1zHa5z8WoF0Fm2VIHoWO_CA0Gaa-f_uy8JGX-NiRO9l2yDaJTxU0nObjG/pubhtml}} shows the raw results.
\begin{table*}[t]
\centering
\small
\resizebox{0.88\textwidth}{!}{
\begin{tabular}{lcccccc}
\toprule
\textbf{Configuration} & \textbf{NQ\textsuperscript{\textdagger}} & \textbf{TriviaQA} & \textbf{PopQA} & \textbf{HotpotQA\textsuperscript{\textdagger}} & \textbf{2Wiki} & \textbf{Musique} \\
\midrule
\rowcolor{orange!20}
\multicolumn{7}{c}{\textbf{{Retrieval: 8, Selection: 3, Turns: 3}}} \\
Full Implementation &{70.5$_\text{(3.2)}$} &{84.0$_\text{(24.6)}$} &{57.7$_\text{(5.9)}$} &{62.4$_\text{(11.1)}$} &{55.1$_\text{(8.3)}$} &26.2$_\text{(7.9)}$  \\
w/o Selection &70.7$_\text{(2.7)}$ &83.1$_\text{(18.0)}$ &57.2$_\text{(8.1)}$ &61.1$_\text{(8.4)}$ &58.9$_\text{(3.3)}$ &22.5$_\text{(1.6)}$ \\
w/o Begin with Search &68.6$_\text{(3.6)}$ &82.2$_\text{(25.5)}$ &55.0$_\text{(7.7)}$ &57.0$_\text{(11.8)}$ &46.8$_\text{(7.9)}$ &20.9$_\text{(2.3)}$ \\
w/o Both &70.8$_\text{(2.5)}$ &83.2$_\text{(18.2)}$ &56.5$_\text{(7.8)}$  &60.1$_\text{(8.7)}$ &57.4$_\text{(3.5)}$ &21.8$_\text{(1.7)}$ \\

\rowcolor{orange!10}
\multicolumn{7}{c}{\textbf{{Retrieval: 5, Selection: 3, Turns: 3}}} \\
Full Implementation &69.6$_\text{(3.5)}$ &83.4$_\text{(24.3)}$ &57.4$_\text{(5.8)}$ &62.0$_\text{(11.9)}$ &53.8$_\text{(7.8)}$ &24.5$_\text{(2.3)}$  \\
w/o Selection &70.8$_\text{(2.6)}$ &81.8$_\text{(19.6)}$ &56.3$_\text{(9.6)}$ &60.8$_\text{(9.4)}$ &57.8$_\text{(3.0)}$ &22.4$_\text{(2.0)}$  \\
w/o Begin with Search &67.6$_\text{(4.0)}$ &81.2$_\text{(26.6)}$ &55.0$_\text{(8.3)}$ &57.4$_\text{(12.0)}$ &50.0$_\text{(9.3)}$ &21.1$_\text{(2.2)}$ \\
w/o Both &70.6$_\text{(2.6)}$ &81.9$_\text{(19.4)}$ &56.0$_\text{(8.6)}$ &60.0$_\text{(9.1)}$ &57.6$_\text{(3.2)}$ &22.3$_\text{(1.7)}$ \\

\rowcolor{orange!5}
\multicolumn{7}{c}{\textbf{{Retrieval: 3, Selection: 3, Turns: 3}}} \\
Full Implementation &69.4$_\text{(3.5)}$ &82.3$_\text{(24.4)}$ &57.0$_\text{(5.7)}$ &61.8$_\text{(11.7)}$ &51.5$_\text{(8.2)}$ &25.1$_\text{(2.3)}$  \\
w/o Selection &69.7$_\text{(5.0)}$ &81.6$_\text{(27.8)}$ &56.1$_\text{(12.5)}$ &59.7$_\text{(11.4)}$ &56.2$_\text{(4.3)}$ &23.5$_\text{(2.2)}$  \\
w/o Begin with Search &67.7$_\text{(3.8)}$ &81.1$_\text{(25.5)}$ &54.2$_\text{(6.7)}$ &58.1$_\text{(11.9)}$ &50.2$_\text{(7.4)}$ &22.1$_\text{(2.5)}$ \\
w/o Both &69.2$_\text{(3.4)}$  &81.5$_\text{(24.4)}$ &55.2$_\text{(8.9)}$ &58.3$_\text{(10.4)}$ &54.6$_\text{(3.3)}$ &22.5$_\text{(2.4)}$ \\

\bottomrule
\end{tabular}}
\caption{Ablation Studies of \texttt{s3} on General Domain RAG. We show generation accuracy as the main results and exact match scores in brackets.}
\label{tab:ablations_general}
\end{table*}

\begin{figure}
\begin{tcolorbox}[colback=orange!4, colframe=brown!90, title=\textbf{Human Evaluation Instruction}]
You are an evaluator for question-answering systems. Your task is to determine whether the system-generated answer aligns with the provided gold (reference) answers.

\textbf{Evaluation Criteria:}
An answer should be marked as \textbf{correct (1)} if it:
\begin{itemize}
    \item Contains the same key information as the golden answers;
    \item Expresses the same meaning, even if using different wording;
    \item Is factually consistent with the golden answers.
\end{itemize}

Please input \textbf{only}:
\begin{itemize}
    \item \texttt{"1"} if the system's answer aligns with the golden answers;
    \item \texttt{"0"} if it does not.
\end{itemize}
\end{tcolorbox}
\caption{Instruction for human evaluation of LLM generation.}
    \label{fig:human_eval_instruct}
\end{figure}

\begin{figure}[!t]
\centering
\includegraphics[width=\linewidth]{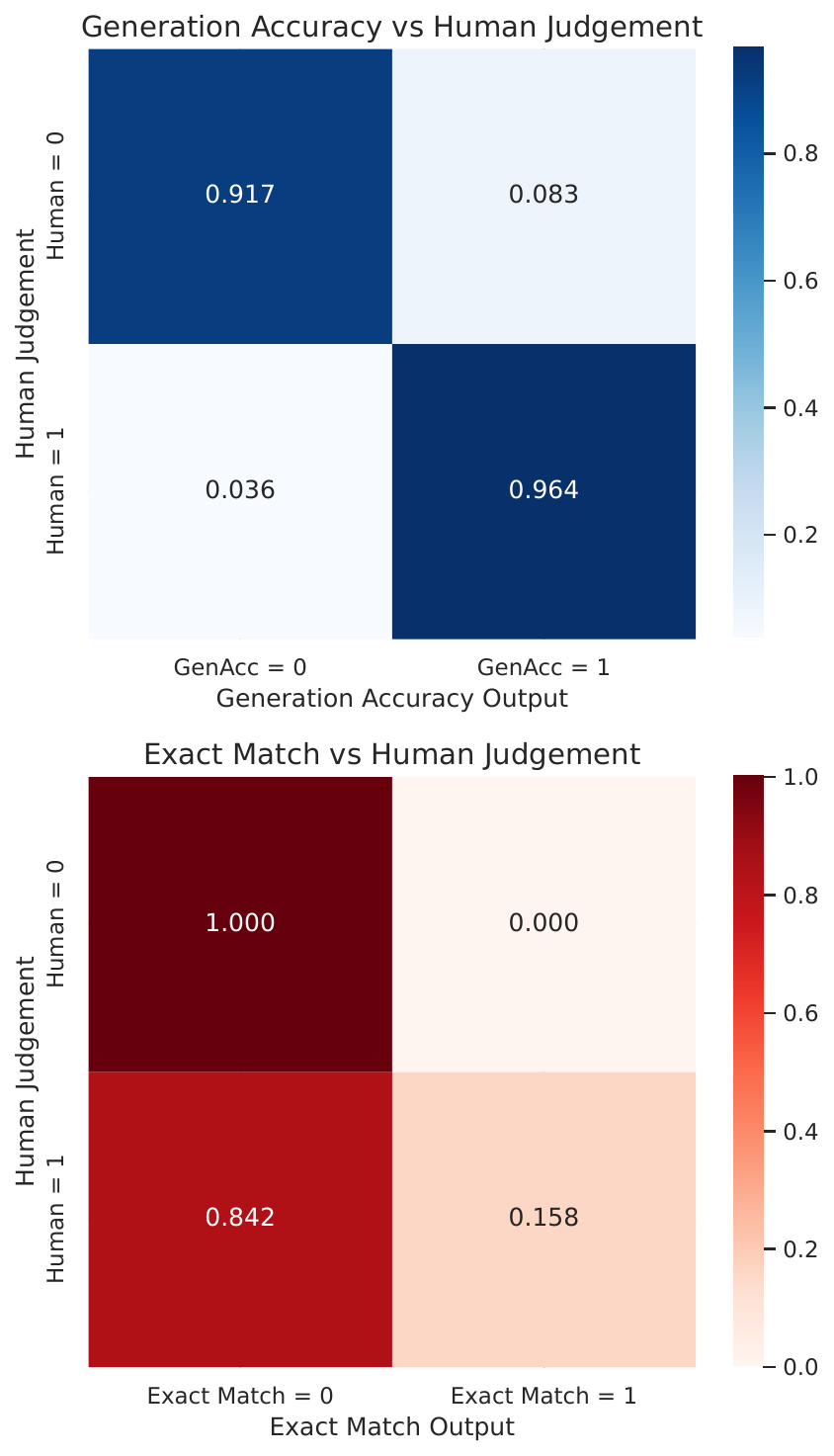}
\caption{Confusion matrices comparing \textbf{Generation Accuracy} (top) and \textbf{Exact Match} (bottom) against human judgment. Each cell indicates the proportion of samples falling into the corresponding category.}
\label{fig:alignment}
\end{figure}

We then compared these human labels against the binary decisions made by \textbf{Generation Accuracy} and \textbf{Exact Match}. As shown in Figure~\ref{fig:alignment}, Generation Accuracy demonstrates strong alignment with human evaluation, correctly identifying 96.4\% of answers that were judged correct by humans. In contrast, Exact Match only captures 15.8\% of such answers, largely due to its strict reliance on string matching.

These results confirm that Generation Accuracy is a more reliable and human-aligned metric, especially for evaluating free-form and abstractive answers where surface forms may differ despite semantic correctness, which also syncs with findings by prior studies applying similar evaluation methods~\cite{song2025r1}.
\begin{figure*}[!t]
\centering
\begin{minipage}[t]{0.48\textwidth}
    \centering
    \includegraphics[width=\linewidth]{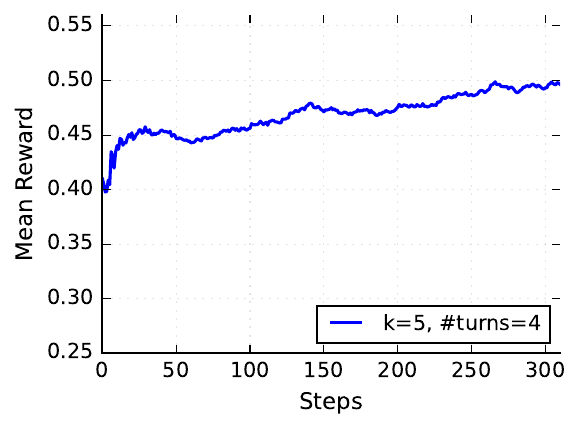}
    \vspace{-1em}
    \caption{Scalability study: mean reward curve when training \texttt{s3} (5-3-4) for 300 steps.}
    \label{fig:scale_study}
\end{minipage}%
\hfill
\begin{minipage}[t]{0.48\textwidth}
    \centering
    \includegraphics[width=\linewidth]{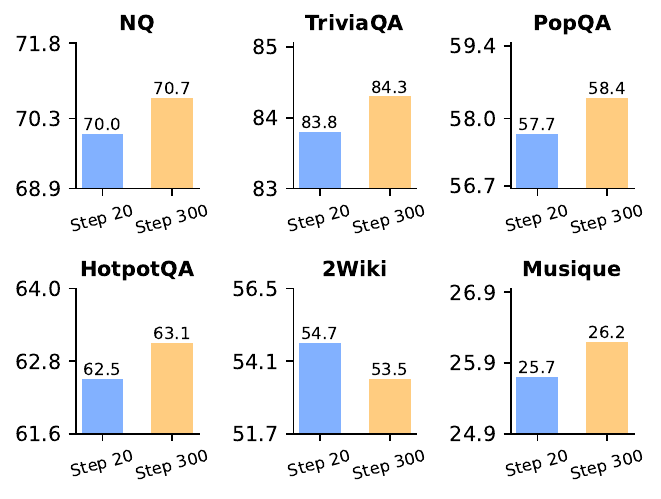}
    \vspace{-1em}
    \caption{Performance comparison at Step 20 vs. Step 300 across datasets.}
    \label{fig:scale_perf}
\end{minipage}
\vspace{-0.5em}
\end{figure*}



\section{Prompts}
\label{ap:prompts}

To train and evaluate the s3 framework effectively, we design three system prompts targeting distinct modules: the search policy (Searcher), answer generation, and judge-based evaluation. Each prompt is carefully constructed to ensure modularity, interpretability, and compatibility with frozen LLMs.

\paragraph{Searcher Prompt.}
The prompt for the Searcher (Figure~\ref{fig:search_prompt}) guides a trained policy to perform structured multi-turn search. It defines a loop-based instruction set that mimics real-world decision-making: the model emits a search query, inspects results, selects key documents, and decides whether to continue searching.
\
This design supports iterative refinement and selection via:
\begin{itemize}[leftmargin=*]
\item \texttt{<query>}: the generated search query in JSON format.
\item \texttt{<information>}: the retrieved documents returned by the search engine.
\item \texttt{<important\_info>}: a subset of documents deemed most relevant (up to 3).
\item \texttt{<search\_complete>}: a binary decision on whether to stop searching.
\end{itemize}
Importantly, only selected documents in \texttt{<important\_info>} are visible to the generator, encouraging the policy to focus on high-quality evidence rather than breadth. By isolating retrieval behavior from generation, this prompt allows reinforcement learning with a frozen black-box LLM using downstream answer quality as a reward.

\paragraph{Answer Generation Prompt.}
Figure~\ref{fig:answer_prompt} shows the prompt used for final answer generation. It provides the accumulated context from selected documents along with the user’s original question. The generator is instructed to produce a direct, succinct answer without verbosity. This format simplifies reward computation and ensures generation outputs are consistent and easy to evaluate.

\paragraph{\texttt{Judge\_Check} Prompt.}
To enable scalable, automated evaluation during training and inference, we employ a lightweight correctness prompt shown in Figure~\ref{fig:answer_check}. This prompt asks an LLM to verify whether any gold answer appears in the predicted response. Unlike brittle exact-match metrics, this approach captures semantically valid completions even if they differ in surface form. During training, a quantized Qwen2.5-14B model is used for cost-effective inference, while evaluation employs Claude-3-Haiku for higher reliability.

Together, these prompts form a coherent pipeline that supports modular training and evaluation of retrieval-augmented generation systems. The clear separation of roles allows s3 to focus learning solely on the search agent, and our prompt designs play a key role in realizing this clean decoupling.

\section{Scalability Study}
\label{sec:scalability}

While \texttt{s3} demonstrates strong performance with just 20 training steps (i.e., 2.4k examples), we investigate how performance evolves with additional data and training. Specifically, we train the ``5-3-4'' configuration for up to 300 steps.

Figure~\ref{fig:scale_study} shows the reward curve over training steps. We observe a consistent upward trend, indicating that the search policy continues to improve with more data and training iterations.

To quantify this improvement, Figure~\ref{fig:scale_perf} compares the model’s QA performance at step 20 and step 300 across six datasets. The results show that \texttt{s3} scales gracefully: most datasets exhibit steady gains, with improvements particularly noticeable on {PopQA}, {HotpotQA}, and {Musique}.

These findings suggest that \texttt{s3} can also benefit from larger-scale training, making it a flexible framework that performs well both in low-resource and high-resource settings.

\clearpage
\begin{figure*}[t]
\centering
\begin{tcolorbox}[title=Prompt Instructions for Search Agent, width=\textwidth, colback=gray!5, colframe=black, fonttitle=\bfseries]
\small
You are a search copilot for a generation model. Based on a user's query and initial searched results, you will first determine if the searched results are enough to produce an answer.
\smallskip

If the searched results are enough, you will use \texttt{<search\_complete>True</search\_complete>} to indicate that you have gathered enough information for the generation model to produce an answer.
\smallskip

If the searched results are not enough, you will go through a loop of \texttt{<query>} $\rightarrow$ \texttt{<information>} $\rightarrow$ \texttt{<important\_info>} $\rightarrow$ \texttt{<search\_complete>} $\rightarrow$ \texttt{<query>} (if not complete) ..., to help the generation model to generate a better answer with more relevant information searched.
\smallskip

You should show the search query between \texttt{<query>} and \texttt{</query>} in JSON format.
\smallskip

Based on the search query, we will return the top searched results between \texttt{<information>} and \texttt{</information>}. You need to put the doc ids of the important documents (up to 3 documents, within the current information window) between \texttt{<important\_info>} and \texttt{</important\_info>} (e.g., \texttt{<important\_info>[1, 4]</important\_info>}).
\smallskip

A search query \textbf{must} be followed by a \texttt{<search\_complete>} tag if the search is not complete.
\smallskip

After reviewing the information, you must decide whether to continue searching with a new query or indicate that the search is complete. If you need more information, use \texttt{<search\_complete>False</search\_complete>}. Otherwise, use \texttt{<search\_complete>True</search\_complete>} to terminate the search.
\smallskip

\textbf{Note:} Only the content between \texttt{<important\_info>} will be used by the generation model to produce an answer.
\smallskip

[An abstractive example of \texttt{s3} loop]

\texttt{<question>}
\{\textcolor{purple}{question}\}
\texttt{</question>}
\smallskip

\texttt{<information>}
\{\textcolor{purple}{initial\_search\_result}\}
\texttt{</information>}

\end{tcolorbox}
\caption{The prompt used for search policy (search agent).}
\label{fig:search_prompt}
\vspace{1em}

\noindent
\begin{minipage}[t]{0.49\textwidth}
\begin{tcolorbox}[title=Prompt for Answer Generation, width=\textwidth, colback=gray!5, colframe=black, fonttitle=\bfseries]
\small
Use the following contexts (some might be irrelevant) on demand:
\smallskip

Contexts:
\{\textcolor{purple}{context}\}
\smallskip

Question: 
\{\textcolor{purple}{question}\}
\smallskip

Important: You MUST directly answer the question without any other text.
\end{tcolorbox}
\captionof{figure}{The prompt used for answer generation by frozen LLM. We enforce the model to direct answer w/o reasoning to avoid the case it includes golden answer in its reasoning chain but not in answer response.}
\label{fig:answer_prompt}
\end{minipage}
\hfill
\begin{minipage}[t]{0.49\textwidth}
\begin{tcolorbox}[title=Prompt for \texttt{Judge\_Check}, width=\textwidth, colback=gray!5, colframe=black, fonttitle=\bfseries]
\small
Please check if any of the golden answers is contained in the following response: 
\smallskip

\{\textcolor{purple}{response}\}
\smallskip

Golden answers: \{\textcolor{purple}{golden\_answers}\}
\smallskip

Please directly answer with 'yes' or 'no'.
\end{tcolorbox}
\captionof{figure}{The prompt used for \texttt{judge\_check}. Qwen2.5-14B-Instruct-GPTQ-Int4 is used in training, and Claude-3-Haiku is used in evaluation.}
\label{fig:answer_check}
\end{minipage}
\end{figure*}

\begin{table*}[!h]
\centering
\small
\renewcommand{\arraystretch}{1.3}
\begin{tabular}{p{3cm} p{11cm}}
\toprule
\textbf{Question} &
\texttt{<question>} What year was the film made that was about the inventor of a type of passive solar house that is made of both natural and upcycled materials such as earth-packed tires? \texttt{</question>} \\
\midrule
\textbf{\texttt{s3} Search Trajectory} &
\texttt{<question>} What year was the film made that was about the inventor of a type of passive solar house that is made of both natural and upcycled materials such as earth-packed tires? \texttt{</question>} \newline

\texttt{<information>} \newline
\textbf{\textcolor{orange}{Doc 1 (Earthship)}}: An Earthship is a type of passive solar house... pioneered by architect Michael Reynolds... \newline
\textbf{Doc 2 (Mike Reynolds)}: Known for Earthship houses using recycled materials like tires... \newline
\textbf{Doc 3 (Don Stephens)}: Discusses earth-integrated designs and upcycled materials... \newline
\texttt{</information>} \newline

\texttt{<search\_complete>} 
\textcolor{red}{\textbf{False}}
\texttt{</search\_complete>} \newline

\texttt{<query>} What year was the film made about \textbf{\textcolor{orange}{the inventor of Earthship}}?" 
\texttt{</query>} \newline

\texttt{<information>} \newline
\textbf{\textcolor{blue}{Doc 1 (Garbage Warrior)}}: Garbage Warrior is a 2007 film about architect Mike Reynolds, inventor of the Earthship style of building... \newline
\textbf{\textcolor{blue}{Doc 2 (Garbage Warrior)}}: A 2007 film following Reynolds and his Earthship building... \newline
{Doc 3 (Earthship)}: Mentions the film Garbage Warrior and its focus on off-grid housing... \newline \texttt{</information>} \newline

\texttt{<important\_info>}\textcolor{blue}{\textbf{[1, 2]}}\texttt{</important\_info>}  \newline

\texttt{<search\_complete>} 
\textcolor{teal}{\textbf{True}}
\texttt{</search\_complete>} \\
\midrule
\textbf{\texttt{s3} Searched Docs} & \textbf{\textcolor{orange}{Doc 1 (Earthship)}}: An Earthship is a type of passive solar house... pioneered by architect Michael Reynolds... \newline
\textbf{Doc 2 (Mike Reynolds)}: Known for Earthship houses using recycled materials like tires... \newline
\textbf{Doc 3 (Don Stephens)}: Discusses earth-integrated designs and upcycled materials... \newline
\textbf{\textcolor{blue}{Doc 4 (Garbage Warrior)}}: Garbage Warrior is a 2007 film about architect Mike Reynolds, inventor of the Earthship style of building... \newline
\textbf{\textcolor{blue}{Doc 5 (Garbage Warrior)}}: A 2007 film following Reynolds and his Earthship building... \\
\midrule
\textbf{Golden Answer} & 2007 \\
\textbf{RAG Answer} & \textit{There is no specific year mentioned for a film made about the inventor of the Earthship, which is a type of passive solar house made of natural and upcycled materials like earth-packed tires. The information provided does not include details about a particular film or its release year.} \\
\textbf{\texttt{s3} Answer} & 2007 \\
\bottomrule
\end{tabular}
\caption{\small An example showing how \texttt{s3} searches and selects correct evidence after issuing a focused search query. RAG fails to answer correctly without temporal grounding.}
\label{tab:case_study_general}
\end{table*}

\end{document}